\documentclass[10pt,twocolumn,letterpaper]{article}

\usepackage{wacv}
\usepackage[accsupp]{axessibility}
\usepackage{times}
\usepackage{epsfig}
\usepackage{graphicx}
\usepackage{amsmath}
\usepackage{amssymb}
\usepackage{array}
\usepackage{pifont}
\newcommand{\cmark}{\ding{51}}%
\newcommand{\xmark}{\ding{55}}%
\usepackage{booktabs}
\usepackage{enumitem, kantlipsum}
\usepackage{authblk}

\newcommand{\var}[1] {(\textit{#1})}

\newcommand{\pname}[0] {SEAM Match-RCNN\xspace}
\newcommand{\pnamenospace}[0] {SEAM Match-RCNN}

\def\wacvPaperID{583} 

\wacvfinalcopy 

\ifwacvfinal
\fi

\ifwacvfinal
\usepackage[breaklinks=true,bookmarks=false]{hyperref}
\else
\usepackage[pagebackref=true,breaklinks=true,colorlinks,bookmarks=false]{hyperref}
\fi

\ifwacvfinal
\else
\fi

\begin{document}
\title{MovingFashion: a Benchmark for the  Video-to-Shop Challenge}

\author[1]{Marco Godi\thanks{indicates equal contribution}}
\newcommand\CoAuthorMark{\footnotemark[\arabic{footnote}]} 
\author[1]{Christian Joppi\protect\CoAuthorMark}
\author[1]{Geri Skenderi\protect\CoAuthorMark}
\author[1,2]{Marco Cristani}
\affil[1]{Department of Computer Science, University of Verona}
\affil[2]{Humatics Srl, Verona, Italy}
\affil[ ]{\tt\small \{marco.godi,christian.joppi,geri.skenderi\}@univr.it}
\affil[ ]{\tt\small marco.cristani@\{univr,humatics\}.it}

\maketitle
\ifwacvfinal\thispagestyle{empty}\fi

\begin{abstract}
   Retrieving clothes that are worn in social media videos (Instagram, TikTok) is the latest frontier of e-fashion, referred to as ``video-to-shop'' in the computer vision literature. In this paper, we present MovingFashion, the first publicly available dataset to cope with this challenge. MovingFashion is composed of 14855 social videos, each one of them associated with e-commerce ``shop'' images where the corresponding clothing items are clearly portrayed. In addition, we present a novel baseline for this scenario, dubbed \pname. The model is trained by image-to-video domain adaptation, allowing the use of video sequences where only their association with a shop image is given, eliminating the need for millions of annotated bounding boxes. \pname builds an embedding, where an attention-based weighted sum of few frames (10) of a social video is enough to individuate the correct product within the first 5 retrieved items in a 14K+ shop element gallery with an accuracy of 80\%. This provides the best performance on MovingFashion, comparing exhaustively against the related state-of-the-art approaches and alternative baselines\footnote{The code for \pname and the MovingFashion dataset are available here: \url{https://github.com/HumaticsLAB/SEAM-Match-RCNN}}. 
\end{abstract}

\section{Introduction}

    One of the most recent challenges in e-fashion is the so-called \emph{video-to-shop}~\cite{cheng2017video2shop,zhaodress}, whose aim is to match a social video (Instagram, TikTok) 
    containing one or more given clothing item(s), against an image gallery, potentially an e-commerce database (Fig.~\ref{fig:MovingFashion}a,b). Individuating the outfit of a celebrity or social influencer can turn videos into priceless commercials, in a market where over a billion hours of video are uploaded and viewed on a daily basis~\cite{duffett2020youtube}.
    Video-to-shop derives from the \emph{street-to-shop} problem, where the probe data is a single image~\cite{hadi2015buy}. On one hand, video-to-shop allows an increase of the available information by adding additional frames as probes. On the other hand, as shown in Fig.~\ref{fig:MovingFashion}b, this information could be noisy due to challenging illumination, drastic zooming, human poses, missing data and multiple people (dis)appearing in the video. Another issue is that a video-to-shop system needs training data with millions of bounding box annotations, linking each box with a shop item~\cite{cheng2017video2shop,zhaodress}.
    
    Our first contribution is MovingFashion, the very first publicly available video-to-shop dataset, composed by $\sim$15K different video sequences, each one related with at least one shop image. 
    The videos of MovingFashion are obtained from the fashion e-shop Net-A-Porter (10132 videos) and the social media platforms Instagram and TikTok (4723 videos), and contain hundreds of frames per shop item, partitioned into a \emph{Regular} and \emph{Hard} setup. 
  
    \begin{table*}[t]
        \centering
        \scriptsize
        \begin{tabular}{l|l l l l l m{16pt} m{16pt} m{26pt} m{17pt} m{24pt}}
            \toprule
             \hfil\textbf{Dataset}&\hfil\textbf{\#Videos}&\hfil\textbf{\#Traject.} &\hfil\textbf{\#FramesXVideo [Avg.]} &\hfil\textbf{\#Shops} &\hfil\textbf{[W, H]} & \textbf{\#Pairs} & \textbf{Wild} & \textbf{Occlusion} & \textbf{Crowd} & \textbf{Available}\\ \hline
            \textit{AsymNet}~\cite{cheng2017video2shop} & \hfil\textit{526} &\hfil\textit{26k} &\hfil\textit{\textbf{n.a.}}&    \hfil \textit{85k} &\hfil\textit{\textbf{n.a.}}& \hfil \textit{39k} & \hfil\textit{\textbf{n.a.}} & \hfil\textit{\textbf{n.a.}} &\hfil\textit{\textbf{n.a.}} &\hfil\textit{\xmark}\\ \hline
            \textit{DPRNet}~\cite{zhaodress} & \hfil\textit{818} &\hfil\textit{5k} & \hfil\textit{\textbf{n.a.}}&\hfil \textit{21k} &\hfil\textit{\textbf{n.a.}}& \hfil\textit{\textbf{n.a.}} & \hfil\textit{\textbf{n.a.}} & \hfil\textit{\textbf{n.a.}} &\hfil\textit{\textbf{n.a.}} &\hfil\xmark\\ \hline\hline
            MovingFashion & \hfil15k & \hfil15k & \hfil390& \hfil14k&\hfil[631$\pm$ 12 , 770$\pm$ 21]& \hfil15k &\hfil\cmark &\hfil\cmark &\hfil\cmark& \hfil\cmark\\ \bottomrule
        \end{tabular}
        \vspace{0.5em}
        \caption{Comparison of Video-2-shop datasets. \emph{n.a.} stands for \emph{not available}.
        }
        \label{tab:dataset}
        \vspace{-1em}
    \end{table*}

    Our second contribution is the SElf-Attention Multiframe (SEAM) Match-RCNN,  a video-to-shop baseline which individuates products and extracts features in a ``street'' video sequence by adopting a feature collection and aggregation mechanism, and then matching the products over a ``shop'' image gallery. SEAM Match-RCNN extends the popular Match-RCNN~\cite{ge2019deepfashion2}, state-of-the-art in the street-to-shop challenge, by applying image-to-video domain adaptation with the use of a novel Multi-frame Matching Head.   
    
    Technically, a pretraining on the image domain of the Match-RCNN enables it to provide initial pseudo-labels for a video sequence, individuating bounding boxes matching a particular product. The training on the target domain exploits our Multi-frame Matching Head, that aggregates features by means of a non-local block~\cite{wang2018non} between different frames, which in turn applies a temporal self-attention mechanism~\cite{gao2018revisiting} and a scoring function. In this way an aggregation based on the attention score is used to create a single descriptor for a clothing item. In practice, \pname allows to train on video data where only the pairs $<$``street'' video,``shop'' image$>$ are available, without annotated ground-truth bounding boxes. This policy permits to alleviate an intense annotation effort, which in the case of MovingFashion would have required drawing $\sim$18M  bounding boxes. In the experiments, \pname gives the best performances on MovingFashion, against multiple baselines and state-of-the-art techniques. Actually, few frames (10) of a social video are enough to individuate the correct product within the first 5 retrieved items with an accuracy of 80\%, making \pname a proof of concept for a potential product in e-fashion.
    Finally, \pname gives explainable results: thanks to self-attention visualization, we understand that the initial quarter of a social video carries the most information for guessing the correct product.


\section{Related Work}

    \textbf{Video-to-Shop.}
    The first street-to-shop approaches employed single street images~\cite{ge2019deepfashion2,hadi2015buy,liu2016deepfashion}; ``street'' video queries followed afterwards, paving the way for video-to-shop methods~\cite{cheng2017video2shop,zhaodress}. AsymNet~\cite{cheng2017video2shop} aggregates frames by exploiting temporal continuity; it combines an LSTM and a binary tree, with each component requiring a separate training procedure. On the contrary, our \pname uses self-attention to learn a descriptor from a bunch of heterogeneous images, where temporal continuity is not required. 
    DPRNet~\cite{zhaodress} manages the video-to-shop problem by treating it as street-to-shop, with a network that detects and tracks garments in the video, selecting automatically the frame with the highest quality (in terms of occlusions, blurring etc). That detection is finally fed into an image-to-image retrieval module. \pname does not perform this kind of tracking, which could be prohibitive on social videos that have strong heterogeneous variations on few frames.
    
    Video-to-shop approaches shares similarities with video person Re-ID~\cite{liu2019spatially}, where the goal is to match a video snippet of a person's silhouette against a gallery of image identities taken from a different camera. 
    State-of-the-art approaches are VKD~\cite{porrello2020robust}, NVAN~\cite{liu2019spatially} and MGH~\cite{Yan_2020_CVPR}. VKD proposes to learn using diverse views of the same target with a teacher-student framework, where the teacher educates a student who observes fewer views. NVAN is based on a non-local block self-attention module, embedded into the backbone CNN at multiple feature levels to incorporate both spatial and temporal characteristics of the pedestrian videos into the representation. Multi-Granular Hypergraph (MGH) is a novel graph-based framework which uses graph networks to cope with this problem.

    \textbf{Video-to-shop datasets.} 
    Unfortunately, no video-to-shop datasets are publicly available. The above quoted \cite{cheng2017video2shop} and \cite{zhaodress} use proprietary datasets, which have been not made open to the scientific community. We compare these datasets and their reported characteristics with our MovingFashion dataset in Table~\ref{tab:dataset}. It is visible that the datasets from AsymNet and DPRNet have a moderate number of sequences (526 and 818, respectively), while MovingFashion contains almost thirty times that amount (15K). In order to create more query data, DPRNet and AsymNet sample multiple sequences from the videos (generating 26K and 5K sub-trajectories, respectively). AsymNet contains 39K exact street-shop pairs and 85K diverse shop items, so one may infer shop distractors are present (shop items not present in the street set) but no details are provided on this. DPRNet has 21K Shop items, with no mentions about the exact pairs. MovingFashion has a single item associated with a unique shop image for each video, for a total of 15K unique (video) street-shop pairs.
    
      \begin{figure}[t!]
            \centering
            \includegraphics[width=\linewidth]{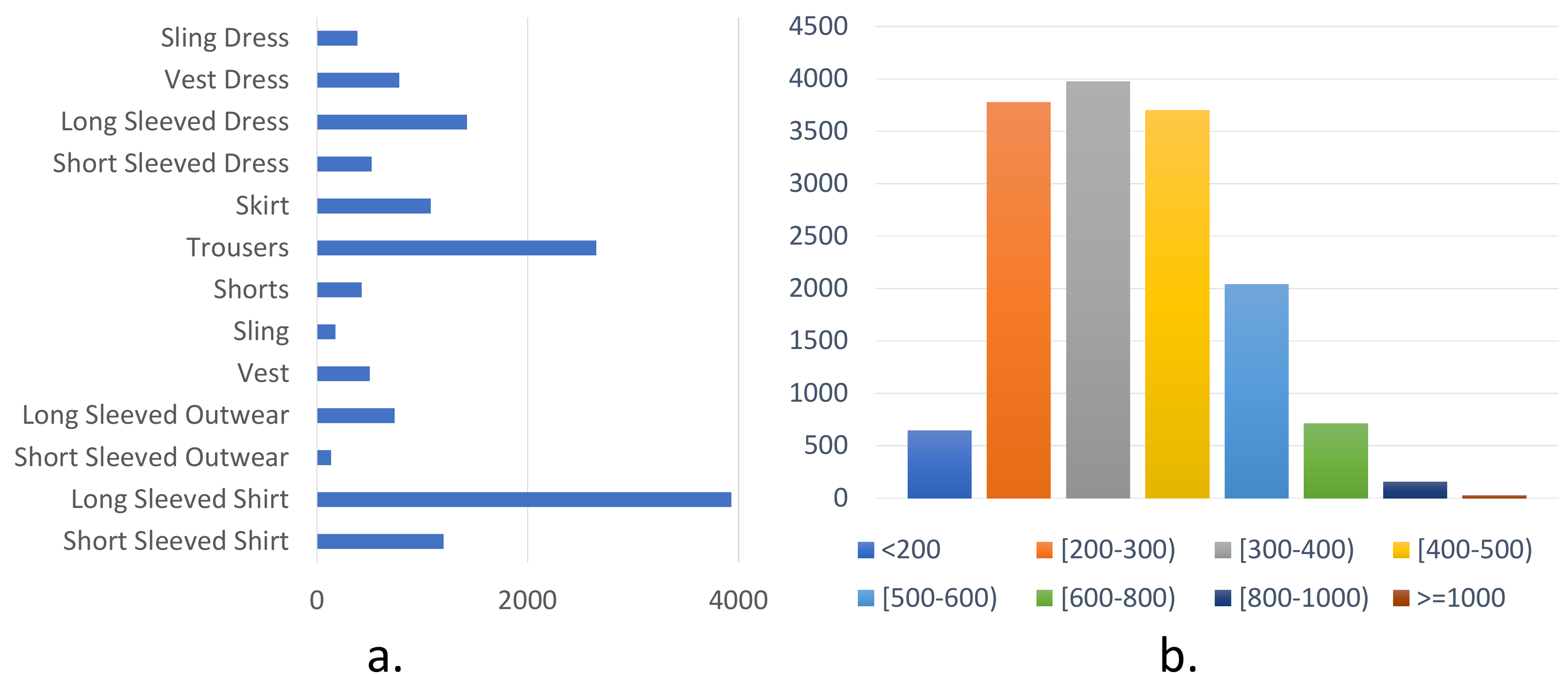}
            \caption{
            MovingFashion statistics; 
            a) Cardinality of each clothing item class;
            b) Histogram of the number of frames for the street sequences.
            }
            \label{fig:MovingFashionStats}
            \vspace{-1em}
    \end{figure}
    The DeepFashion2 dataset (DF2)~\cite{ge2019deepfashion2} presents a particular scenario:  
    DF2 is made for the street-to-shop challenge, but some shop items are related to more than one street image (coming from different sources), creating 11K pairings. This provides us with another experimental setting. 

\section{MovingFashion}\label{sec:dataset}
    \begin{figure}[t!]
        \centering
        \includegraphics[width=\linewidth]{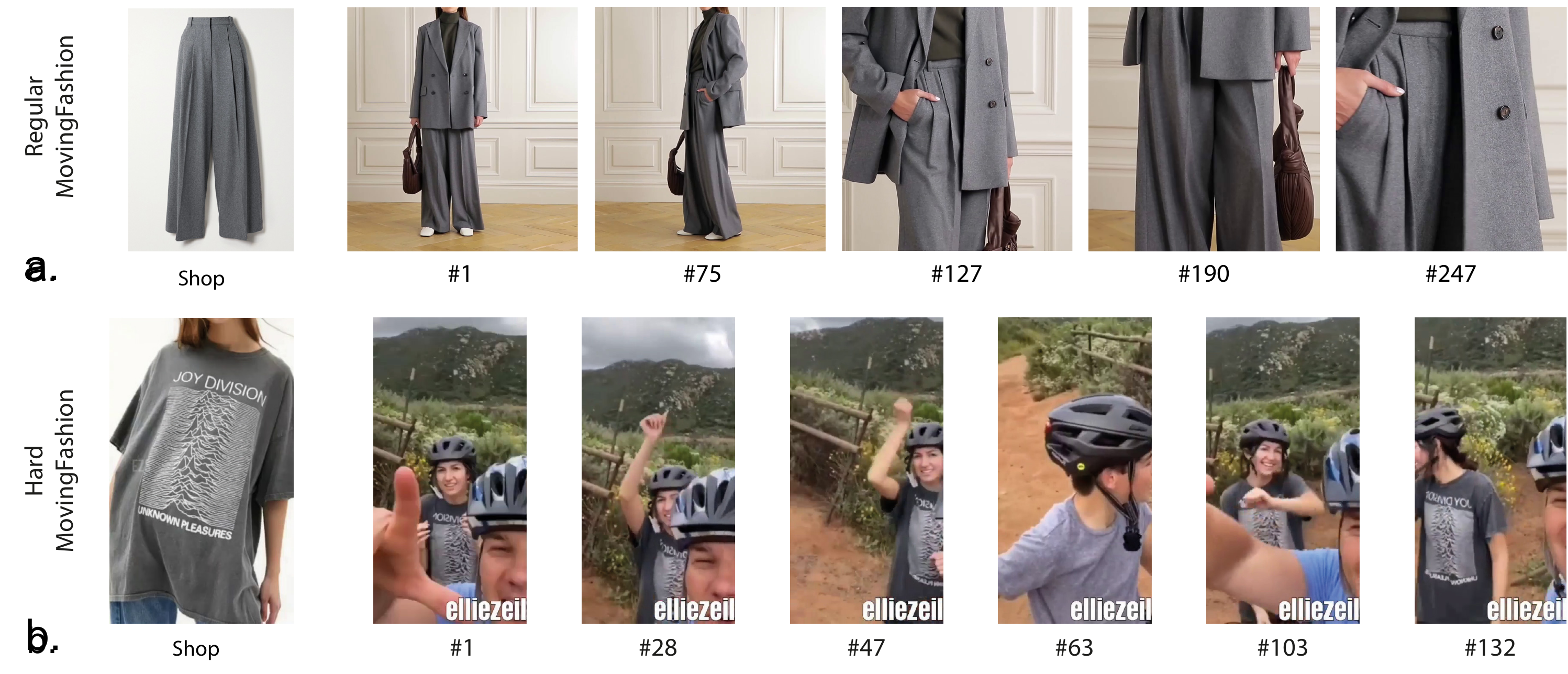}
        \caption{MovingFashion dataset samples. The top row contains a ``Regular'' sequence, the bottom row a ``Hard'' sequence.
        }
        \label{fig:MovingFashion}
    \end{figure}
    MovingFashion has 5.854M annotated frames, organized into 15045 video-shop \emph{matching pairs}, i.e., each video is associated with a distinct \emph{shop image}. In particular, there are 14.8K \emph{unique} videos, among which some sequences (190 videos) have more than one associated shop item (\emph{e.g.}, a t-shirt and trousers). The length of the videos is detailed in Fig.~\ref{fig:MovingFashionStats}b, while the frame rate amounts to about 30FPS. 
    Shop items are divided in classes, following the DeepFashion2~\cite{ge2019deepfashion2} taxonomy. The list of classes and the number of occurrences for each class in the dataset is reported in Fig.~\ref{fig:MovingFashionStats}a.
   
    \subsection{Data sources}
        MovingFashion is formed by two subsets: \emph{Regular} and \emph{Hard}.
        \\
        \textbf{Regular MovingFashion}: 
            Regular MovingFashion consists of 10132 videos downloaded from the e-commerce website Net-A-Porter~\footnote{\url{https://www.net-a-porter.com}}: in the street video a single person is wearing the shop item in an indoor scenario (which can vary), and the corresponding shop image consists in the shop item captured over a plain background. This is the canonical shop image we have used in our experiments. Additionally, we have collected: a \emph{front} shop image captured in the same background of the street video and worn by the same model in a frontal pose; a \emph{rear} view image and a detail of the \emph{fabric}. These last three were not used in the experiments. An example of Regular MovingFashion is showed in Fig.~\ref{fig:MovingFashion}a (more in the supplementary material).
        \\    
        \textbf{Hard MovingFashion:} Hard MovingFashion consists of 4723 videos from the social platforms Instagram and TikTok. In this case, shop images have been obtained either by downloading images associated to the video as multiple images of the Instagram post or as part of the video itself (the spatial layout of some raw videos was organized in two halves, one being a still
            picture of the “shop” item, the other with the “street” video). Hard MovingFashion represents the hardest challenge, since all of the critical conditions listed in the introduction are present here, as also visible in Fig.~\ref{fig:MovingFashion}b.

    \subsection{Data Collection and cleaning}\label{sec:datacoll}
        All of the videos in Net-A-Porter have been designed to promote a clothing item, which made the data collection process simpler. Cleaning was necessary only to remove classes not compliant with the taxonomy of DeepFashion2~\cite{ge2019deepfashion2}. In contrast, Instagram and TikTok videos required a lot of work, starting with the search for the street videos and their shop counterparts using the available API, up to the careful scraping of hashtags and profiles. Other minor but time-consuming issues (fully/partially duplicate videos, wrongly associated shop items) are discussed in the supplementary material. 
        
        After collection and cleaning, we split the data into a training and a testing partition, taking care of applying the same split for each single class. We perform a 90/10 train/test split. Bounding boxes are extracted using a clothing detector. We then utilize the training data to train our \pname following the unsupervised procedure shown in Sec.~\ref{sec:unsuptrain}. Since video sequences contain more than one item, to evaluate \pname and all the comparative approaches we create a tracklet containing the correct item for each street video sequence. That is done by selecting the tracklet that matches most with the shop item, following the \emph{training tracking procedure} detailed in Sec.~\ref{sec:unsuptrain}. A \emph{tracklet} is a set of consecutive detections which refer to the same object. We manually check each one of these to ensure that at least 50\% of the detections in the tracklet actually contain the shop item. For the detections which are too noisy (i.e. they do not focus on any precise clothing item), we dropped the entire sequence, in order to speed up the collection procedure. Fortunately, this happened on a minority of sequences ($\sim$150 videos), indicating a general success of the tracking procedure. The remaining tracklets are kept as noisy annotations in JSON format. All of the comparative approaches shown in Sec.~\ref{sec:exp} use these \emph{ground truth tracklets} for training and testing.
    
     \begin{figure*}[t!]
                \centering
                \includegraphics[width=\linewidth]{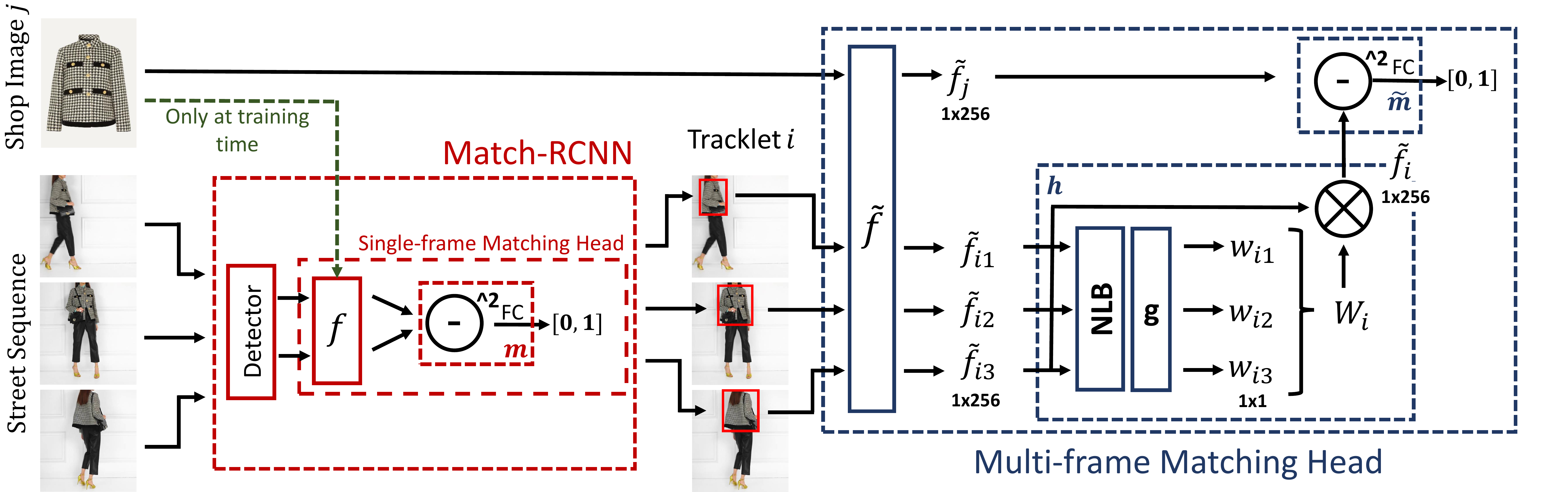}
                \caption{
                Architecture of our \pname system. Images are first processed by the Match-RCNN to extract bounding boxes and convolutional features. After tracking a clothing item across frames, its features are further processed by the Multi-frame Matching Head producing a final matching score between the street video sequence and the shop image.
                }
                \label{fig:seam_rcmm}
                \vspace{-1em}
        \end{figure*}
\section{\pname}\label{sec:method}
    \pname takes as input a sequence of street images  ${i_1 ... i_N}$, and compares it with the gallery of $K$ shop images providing a list of matching scores as output. Once the model has learned, the retrieval happens by means of three procedures: 1) \emph{Tracklet creation}; 2) \emph{Feature aggregation}; 3) \emph{Video-to-shop matching}.
    Going through these steps will allow us to present the structure of the network, detailed in Fig.~\ref{fig:seam_rcmm}.
    
    \subsection{Tracklet creation}\label{sec:trackletextr}
    On the input video sequence we need to locate a set of consecutive detections which refer to the same object, dubbed here \emph{tracklet}. Since multiple objects might be on the video, multiple tracklets are expected.  The module that deals with this is the \textbf{Match-RCNN}, which is composed of three functions:
    \begin{enumerate}[noitemsep]
        \item A clothing detector which provides convolutional features $c_{i,t,k}$ with $i$ indicating the $i$-th tracklet, $t$ indicating the frame, $k$ the $k$-th detection in that frame;
        \item A 256-d feature extractor $f_{i,t,k}=f(c_{i,t,k}) \in \mathbb{R} ^{256}$ which performs embedding of the convolutional features;
        \item A matching score function $m(f_{i,t,k},f_{i,t',k'}) \in [0,1]$, comparing different embeddings.
    \end{enumerate}
    $f$ and $m$ together form the \textbf{Single-frame Matching Head}.

    The tracklet extraction procedure is performed in an iterative fashion, following a two-step process:
    \begin{enumerate}[noitemsep]
        \item Determining the \emph{pivot} bounding box: This represents the most confident detection $f_{i,t_{best},k_{best}}$ in the sequence and acts as the central reference based on which the tracklet will be built.
        \item Performing \emph{propagation} based on the \emph{pivot}: By comparing the embedding of the pivot $f_{i,t_{best},k_{best}}$ with all of the detections in every frame, the tracklet $i$ can be built. In particular, a detection joins the tracklet if its matching score (matching function $m$ of the Single-Frame Matching Head) is above a certain threshold, to avoid considering frames where the item is not visible. 
    \end{enumerate}
    Once the tracklet $i$ is built, its detections are removed, and another tracklet focusing on a different item can be built.
    

    \subsection{Feature aggregation}\label{sec:featuresextraction}
    The next step is aggregating the information of a tracklet and condensing it into a single multi-frame descriptor. The module that deals with the feature aggregation procedure is the \textbf{Multi-frame Matching Head} and it is composed of the following functions and modules:
    \begin{enumerate}[noitemsep]
        \item A 256-d feature extractor $\tilde{f}_{i,t}=\tilde{f}(c_{i,t}) \in \mathbb{R} ^{256}$ operating on the bounding box at time $t$ of the tracklet $i$, i.e., $c_{i,t}$.
        %
        
        \item A non-local block~\cite{wang2018non} module which applies self-attention, enriching $\{\tilde{f}_{i,t}\}_t$ with information coming from all the other bounding boxes related to the object tracklet $i$.
        \item An attention module  $g\colon \mathbb{R}^{N \times 256} \mapsto \mathbb{R}^{N}$ that for each detection in a tracklet computes an importance score $w_t$ such that $\sum_{t} w_t = 1$. 
        \item An aggregation module, which fuses $\{\tilde{f}_{i,t}\}_t$ into a joint descriptor $\tilde{f}_{i}$ 
        by a weighted average over the attention scores $\{w_t\}$: $h(x)=g(NLB(x)) \cdot x$, $x \in \mathbb{R}^{N \times 256}$. 
        \item A matching score function $\tilde{m}(\tilde{f}_{j}, \tilde{f}_{i}) \in [0,1]$, which compares the aggregated descriptor for item $k$ and street sequence $i$ ($h(\{\tilde{f}_{i,t}\}_t)$ as $\tilde{f}_{i}$) with the the shop descriptor of image $j$.
    \end{enumerate}
    
    The aggregation procedure starts with the feature extractor $\tilde{f}$, which creates the initial descriptors for each box in a sequence. Then, self-attention is computed by the non-local block module and afterwards the attention module calculates the attention weights for each descriptor. The aggregation module puts all of the above together, producing the single multi-frame descriptor $\tilde{f}_{i}$. Note that we discard temporal continuity \emph{by design}. Social network videos usually have dramatic zooms, very fast pose dynamics and occlusions; moreover, elaborated videos may have shot changes which can fragment temporal continuity. 

    \subsection{Video-to-shop matching}\label{sec:matching}
    Following the feature aggregation procedure described above, we obtain a single multi-frame descriptor $\tilde{f}_{i}$ of the street tracklet $i$. In this final procedure, the matching score function $\tilde{m}$ of the Multi-frame Matching Head is used to match the aggregated multi-frame description with the single 
    shop descriptor of image $j$, $\tilde{f}_{j}$ (which can be considered as a tracklet composed by a single frame), under the assumption that a single item is portrayed in the shop image. 
    We use the matching function $\tilde{m}$ on all the images in the shop gallery, producing in this way a list of matching scores between the street tracklet and all the images in the shop gallery, sorted in descending order, creating thus a \emph{ranking}.
    

    \subsection{Model Training}\label{sec:unsuptrain}
        To avoid the need of bounding boxes annotations, a time-consuming procedure especially for videos, \pname is trained by domain adaptation, through two phases: pretraining on the source image domain and training on the target video domain. 
        
        \textbf{Pretraining on Source domain}.
        The Match-RCNN part of \pname (detector and Single-frame Matching Head) is pretrained on an image street-to-shop dataset (e.g. DeepFashion2). 
        The purpose of this phase is to train a model that is able to estimate bounding boxes and matching scores in the target domain (even with noisy predictions due to the domain gap). Such predictions are used to generate tracklets and pseudo-labels to train the Multi-frame Matching Head.

        \textbf{Training on Target domain}.
        The training procedure estimates the parameters for the Multi-frame Matching Head of the \pname, whose structure is detailed in Sec.~\ref{sec:featuresextraction}, and fine-tunes the Single-frame Matching Head, while the detector's weights are frozen. The weights of the features extractor $\tilde{f}$ and matching score function $\tilde{m}$ are initialized copying those of $f$ and $m$ from the pretrained Single-frame Matching Head. Conversely, the attention modules of $h$ are randomly initialized. 
        During training, image and street video sequence pairs (thanks to the MovingFashion pairing annotations) are sampled, which are leveraged in the tracking procedure (Sec.~\ref{sec:trackletextr}): the pivot selection is done by selecting the detection that matches the shop product the most in the whole video if the matching score inferred from the matching function $m$ of the Single-frame Matching Head is over a certain threshold. The propagation step remains the same as in Sec.~\ref{sec:trackletextr}.
        With this \emph{training tracking procedure} a tracklet is built such that, with a certain confidence, it contains the correct shop item due to the pivot selection starting from the ground truth shop image. This is considered as a positive match during training (i.e. we set 1 as a pseudo-label for the tracklet). For what concerns the Single-frame Matching Head fine-tuning, each detection that composes the tracklet is considered as positive match as well.
        The tracklet is then passed as input to the Multi-frame Matching head, which computes a singular multi-frame descriptor $\tilde{f}_{i}$ thanks to the aggregation procedure described in Sec.~\ref{sec:featuresextraction}. In the end, this singular multi-frame descriptor $\tilde{f}_{i}$ is compared with the corresponding shop descriptor $\tilde{f}_{j}$ (obtained by using the feature extractor $\tilde{f}_{j}=f(c_{j})$) utilizing the matching score function $\tilde{m}$. This produces a matching score in the range [0, 1]. 
        
        Training is done by Stochastic Gradient Descent using cross-entropy loss for the classification of street videos and shop images as positive/negative matches. Positive pairings are built using the aforementioned procedure. All of the other combinations between tracklets and shop image descriptors are considered negative pairings (i.e. pseudo-label of 0) for the Single-frame Matching Head and the Multi-frame Matching Head.

           \begin{table*}
        \centering
        \scriptsize
        \begin{tabular}{l|l l l l | m{14pt} m{14pt} m{16pt} m{16pt}| m{14pt} m{14pt} m{16pt} m{16pt}}
            \toprule
            \textbf{Method} & \multicolumn{4}{l|}{\hfil\textbf{MovingFashion}} & \multicolumn{4}{l|}{\hfil \textbf{Regular-MovingFashion}}& \multicolumn{4}{l}{\hfil \textbf{Hard-MovingFashion}}\\
             & \hfil\textbf{T-1} & \hfil\textbf{T-5} & \hfil\textbf{T-10} & \hfil\textbf{T-20} & \hfil\textbf{T-1} & \hfil\textbf{T-5} & \hfil\textbf{T-10} & \hfil\textbf{T-20} & \hfil\textbf{T-1} & \hfil\textbf{T-5} & \hfil\textbf{T-10} & \hfil\textbf{T-20}\\ \hline

            Max Confidence~\cite{ge2019deepfashion2}     & 0.29\var{0.022}      &  0.59\var{0.020} &  0.72\var{0.018}&  0.83\var{0.017} & 0.31      &  0.63 &  0.76  &  0.86 & 0.21      &  0.46 &  0.60  &  0.71\\ \hline
            Max Matching~\cite{cheng2017video2shop}    & 0.26\var{0.025}      &  0.60\var{0.022} &  0.74\var{0.021}  &  0.85\var{0.016} & 0.29      &  0.65 &  0.79  &  0.88& 0.17      &  0.44 &  0.58  &  0.73\\ \hline
            \hline
             NVAN (2019)~\cite{liu2019spatially}&       0.38\var{0.023}      &  0.62\var{0.021}   &   0.70\var{0.022}   &    0.80\var{0.021}   & 0.47      &  0.73 &  0.81  &  0.90 & 0.11    &  0.28 &  0.37  &  0.48\\ \hline
            VKD (2020)~\cite{porrello2020robust}&       0.40\var{0.019}    & 0.49\var{0.019} &    0.58\var{0.018} &    0.65\var{0.019} & 0.49     &  0.59 &  0.68  & 0.75 & 0.13 & 0.20 &  0.27  &  0.34\\ \hline
        MGH (2020)~\cite{Yan_2020_CVPR}&       0.40\var{0.021}      &  0.59\var{0.020}   &   0.66\var{0.020}   &    0.74\var{0.019}   & 0.47      &  0.67 &  0.73  &  0.81 & 0.18    &  0.35 &  0.43  &  0.52\\ \hline\hline
            AsymNet (2017)~\cite{cheng2017video2shop}&       0.42\var{0.023}    & 0.73\var{0.019} &    0.86\var{0.016} &    0.92\var{0.011} & 0.49     &  0.81 &  0.93   &  0.96  & 0.22     &  0.47 &  0.65  &  0.74\\ \hline
            AsymNet [AVG]&      0.39\var{0.023}    &  0.66\var{0.022} &   0.83\var{0.019} &   0.90\var{0.015}& 0.46      &  0.78 &  0.90  &  0.96 & 0.19    &  0.44 &  0.62  &  0.73\\ \hline
            AsymNet [MAX]&      0.40\var{0.021}   &  0.71\var{0.020} &    0.81\var{0.017} &    0.90\var{0.014} & 0.47      &  0.80 &  0.91  &  0.95 & 0.20      &  0.42 &  0.61  &  0.73\\ \hline\hline
             Average Distance~\cite{cheng2017video2shop}   & 0.39\var{0.021} &  0.73\var{0.020} &  0.84\var{0.017}  &  0.91\var{0.013} &  0.43      &  0.79 &  0.88  &  0.94 & 0.24      &  0.56 &  0.70  &  0.81\\ \hline
             \textbf{SEAM M-RCNN w/o NLB, $g$}    & 0.37\var{0.031}      &  0.73\var{0.025} &  0.86\var{0.020}  &  0.93\var{0.015} & 0.42 &  0.78 &  0.90  &  0.95 & 0.21 &  0.57 &  0.75  &  0.85 \\ \hline
              \textbf{SEAM M-RCNN w/o NLB}    & 0.41\var{0.021}      &  0.73\var{0.018} &  0.83\var{0.015} &  0.91\var{0.012} & 0.47      &  0.79 &  0.89  &  0.95& 0.21      &  0.54 &  0.66  & 0.79\\ \hline
            \textbf{SEAM M-RCNN}    & \textbf{0.49\var{0.022}}      &  \textbf{0.80\var{0.018}} &  \textbf{0.89\var{0.016}} &  \textbf{0.94\var{0.010}} & \textbf{0.55}      &  \textbf{0.86} &  \textbf{0.94}  &  \textbf{0.97}& \textbf{0.30}      &  \textbf{0.62} &  \textbf{0.76}  &  \textbf{0.87}\\ \bottomrule
            
        \end{tabular}
        \vspace{0.5em}
        \caption{\label{tab:results} Video-to-Shop retrieval results on MovingFashion.
        Note: T-K means Top-K Accuracy.}
        \vspace{-1em}
        \end{table*}
\section{Experiments}\label{sec:exp}

    For the retrieval performance evaluation, we follow the testing protocol of DeepFashion2~\cite{ge2019deepfashion2} for evaluating a street image probe against a shop image gallery, with some modifications in order to cope with videos. In DeepFashion2, a street image generates multiple detections: each \emph{street} detection can generate a \emph{proper} matching with some shop image, if it overlaps by a threshold with the corresponding ground truth street bounding box and if its item class is correct, otherwise the matching score is 0.
    
    On MovingFashion,  we compute detections on every street image and we build tracklets using the \emph{tracking procedure} detailed in Sec.~\ref{sec:trackletextr}.
    Then, we compute the average IoU between each street tracklet and the \emph{ground truth tracklet}. The one with the highest average IoU is chosen and used as a query. 
    In order to guarantee fairness in experiments, all baselines  and  comparative  methods  have  been pretrained on two different street-to-shop datasets: DeepFashion2 and Exact Street2Shop~\cite{hadi2015buy}; the former has 873K probe-gallery pairs, while the latter 39K pairs only. Detailed results are reported for the first case, since performances were higher, while in the second case we show the main retrieval results, where our SEAM Match-RCNN remains the best performing approach.

        \begin{table}[h!]
            \centering
            \footnotesize
            \begin{tabular}{|m{95pt}| m{20pt} m{20pt} m{20pt} m{20pt} |}
            \hline
                \textbf{Method} & \textbf{T-1} & \textbf{T-5} & \textbf{T-10} & \textbf{T-20} \\ \hline
                NVAN~\cite{liu2019spatially} & 0.07 & 0.20 & 0.29& 0.42 \\ \hline
                VKD~\cite{porrello2020robust} & 0.16 &0.24 &0.31 &0.38 \\ \hline
                MGH~\cite{Yan_2020_CVPR} & 0.15 &0.23 &0.30 &0.41 \\ \hline
                AsymNet~\cite{cheng2017video2shop} & 0.09& 0.26& 0.37& 0.49 \\ \hline
                \textbf{\pname} & \textbf{0.21} & \textbf{0.41}& \textbf{0.53}& \textbf{0.62}\\ \hline
                
            \end{tabular}
            \vspace{0.5em}
            \caption{\label{tab:s2s} Top-K accuracy on MovingFashion, pretraining on S2S~\cite{hadi2015buy}}
            \vspace{-1em}
        \end{table}


    \subsection{Experiments on MovingFashion}\label{sec:expmf}
    
        We compare our technique with three types of approaches:

            \textbf{Multi-frame baselines}. They are extensions of single-frame techniques to multi-frame. The \emph{Max Confidence}~\cite{ge2019deepfashion2} keeps the most confident detection in a tracklet and uses it for Single-frame Matching, employing a Match-RCNN. The \emph{Max Matching} is inspired from~\cite{cheng2017video2shop} and considers the max matching score between the tracklet's street frames and each shop image. These two baselines are actually working with a single image, which is selected by looking at the entire pool of frames in a tracklet. They are also useful to validate the performance boost that comes when using multiple frames instead of single ones.
            
            The \emph{Average Distance} is inspired by ~\cite{cheng2017video2shop} and consists in averaging single-image matching scores of the tracklet street frames and each shop image.
            The \emph{\pname w/o $NLB$,$g$} is obtained by averaging \emph{descriptors} (and not matching scores) together by average pooling, removing in practice the NLB self-attention block and the attention scoring function $g$ from the \pname (see the scheme in Fig.~\ref{fig:seam_rcmm}). Finally, \emph{\pname w/o $NLB$} keeps the attention score, without the self-attention. These last three are proper multi-frame baselines, in the sense that they merge information coming from multiple frames.
            
            \textbf{Video Re-Identification approaches}. Video Re-Id approaches share similarities with Video-to-shop, in that they look for the best way to aggregate multi-frame information to match a person in a disjoint multi-camera setting. In practice, we consider each shop clothing item the equivalent of a Person Re-Identification Identity. The main differences between video-to-shop and Person Re-ID are that in our case less information is available in terms of pixels, since
            face and hair need to be discarded, focusing only on the clothing. 
            Here we consider the SoA approaches of NVAN~\cite{liu2019spatially}, VKD~\cite{porrello2020robust} and MGH~\cite{Yan_2020_CVPR}\footnote{At the moment of writing, the MGH approach is state-of-the-art in the MARS Video Person Re-Identification dataset, followed closely by VKD and NVAN.}. 
        
            \textbf{Video-to-shop approaches}. We consider the \emph{AsymNet}~\cite{cheng2017video2shop} approach\footnote{The code is available at \url{https://github.com/kyusbok/Video2ShopExactMatching}.}, and its modifications AsymNet[AVG] and AsymNet[MAX], in which the aggregations are made respectively by the average and the max of the similarity score nodes' outputs instead of using the fusion nodes binary tree. Asymnet exploits temporal continuity, yet it does not reach our results.
        
    We set the sequence length to $T=10$ for both training and testing, picking the frames using the Restricted Random Sampling strategy~\cite{li2018diversity}, thus ensuring coverage of the entire sequence length. To analyze variability in the results, we analyze the testing samples by sub-sampling them into pool of 800, 20 times, averaging the rankings and computing standard deviations.

    Table~\ref{tab:results} reports the results. Three facts become apparent: 1) As expected, single-frame approaches (Max Confidence, Max Matching) are definitely inferior ($<$15\% on average) than multi-frame approaches; 2) The considered re-identification approaches, apart from top-1 scores, are inferior to genuine video-to-shop methods; 3) Our \pname surpasses all the competitors, including AsymNet, which gives a better aggregation than the AVG-distance in its [AVG] version and the MAX-distance in its [MAX] version. 
    By looking at the ablative versions of \pname, one can note that the self-attention gives the strongest performance boost, followed by the attention layer. Their cooperation, i.e., the complete \pname, reaches the highest score. 
    
\begin{figure}[t!]
            \centering
            \includegraphics[width=\linewidth]{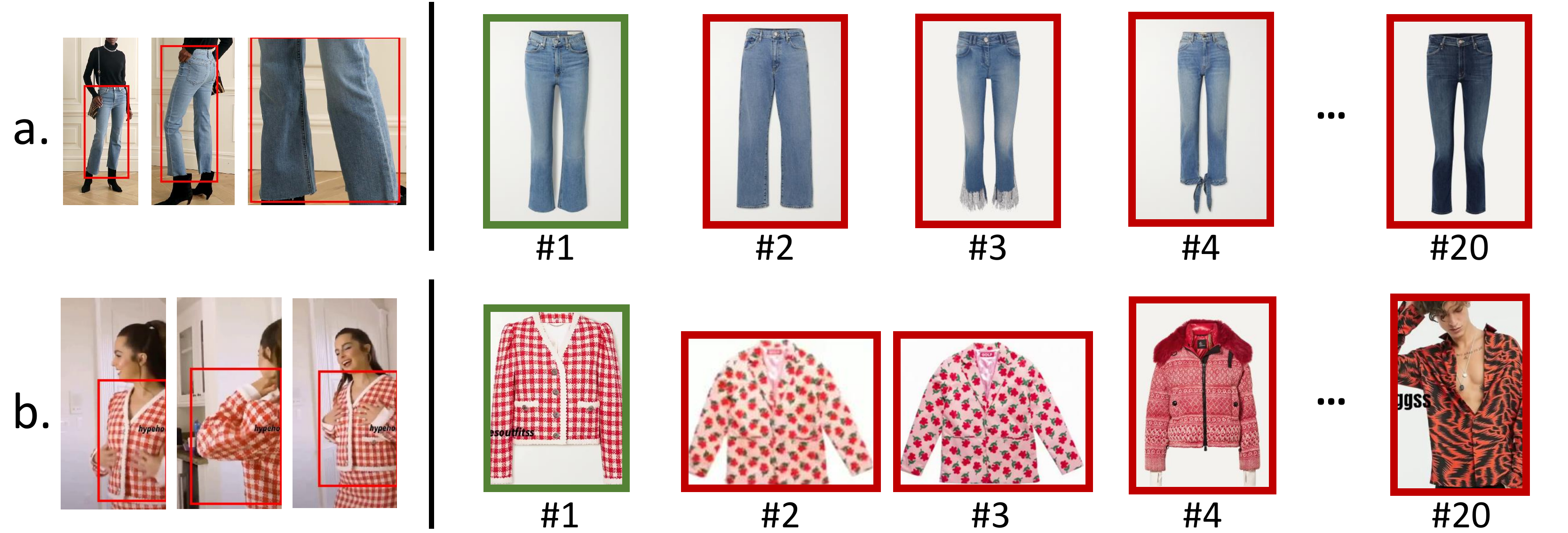}
            \caption{Qualitative retrieval results of \pname for the MovingFashion dataset. On the left, we show 3 frames sampled from the 10 frames used for aggregation. On the right the shop images retrieved with the corresponding rank. The correct matches are with a green border, otherwise red.
            }
            \label{fig:qualitative}
            \vspace{-0.5cm}
        \end{figure}

    By looking at the results within the Regular and Hard MovingFashion partitions, it is quite easy to note the general decrease in performance when it comes to the hard partition. To understand the performance qualitatively, Fig.~\ref{fig:qualitative} shows retrieval results from Regular (Fig.~\ref{fig:qualitative}a) and Hard (Fig.~\ref{fig:qualitative}b). Actually, even if Regular is apparently harder due to many shop alternatives which differ by fine grained results (see the flared jeans), the dramatic changes of poses and backgrounds of the Hard partition play a stronger role. 
    
\begin{figure}[t!]
            \centering
            \includegraphics[width=\linewidth]{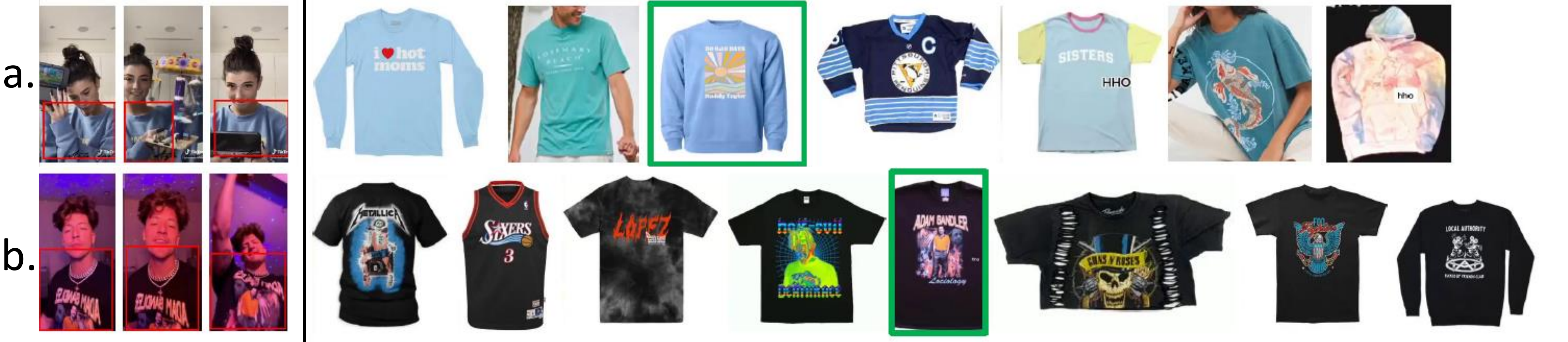}
            \caption{Failure cases results of \pname for the MovingFashion dataset. On the left, we show 3 frames sampled from the 10 frames used for aggregation. On the right the shop images retrieved with the corresponding rank. The correct matches are with a green border.
            }
            \label{fig:failure}
            \vspace{-0.5cm}
        \end{figure}

    Failure cases arise when the original video has discriminant parts of the clothing item covered for most of the sequence, for instance  the logo of the light blue sweatshirt (Fig.~\ref{fig:failure}a). In this case, self-attention overlooks such important details. Complex visual patterns like the hard-rock band logo (Fig.~\ref{fig:failure}b), seem to be not well characterized, meaning that the best match is attributed considering the shape of the logo rather than its content (the “Metallica” logo has the same shape of the probe logo).
    
    The results w.r.t the single clothing classes of MovingFashion are reported in Table~\ref{tab:results_categories}, where it is possible to observe our advantage in all but three classes. Interestingly, we found that the simpler the clothing in terms of texture, the lower the retrieval performance. This is reasonable, since texture adds discriminative details, and this is why classes with simpler texture like vest, sling, shorts and trousers performed worse. We computed textureness by gray-level co-occurrence matrix contrast; quantitatively speaking, textureness and top-1 accuracy in retrieval are found to be correlated (Spearman $\phi=0.72$, $p-value\leq 0.05$).
         
\begin{figure}[t!]
            \centering
            \includegraphics[width=\linewidth]{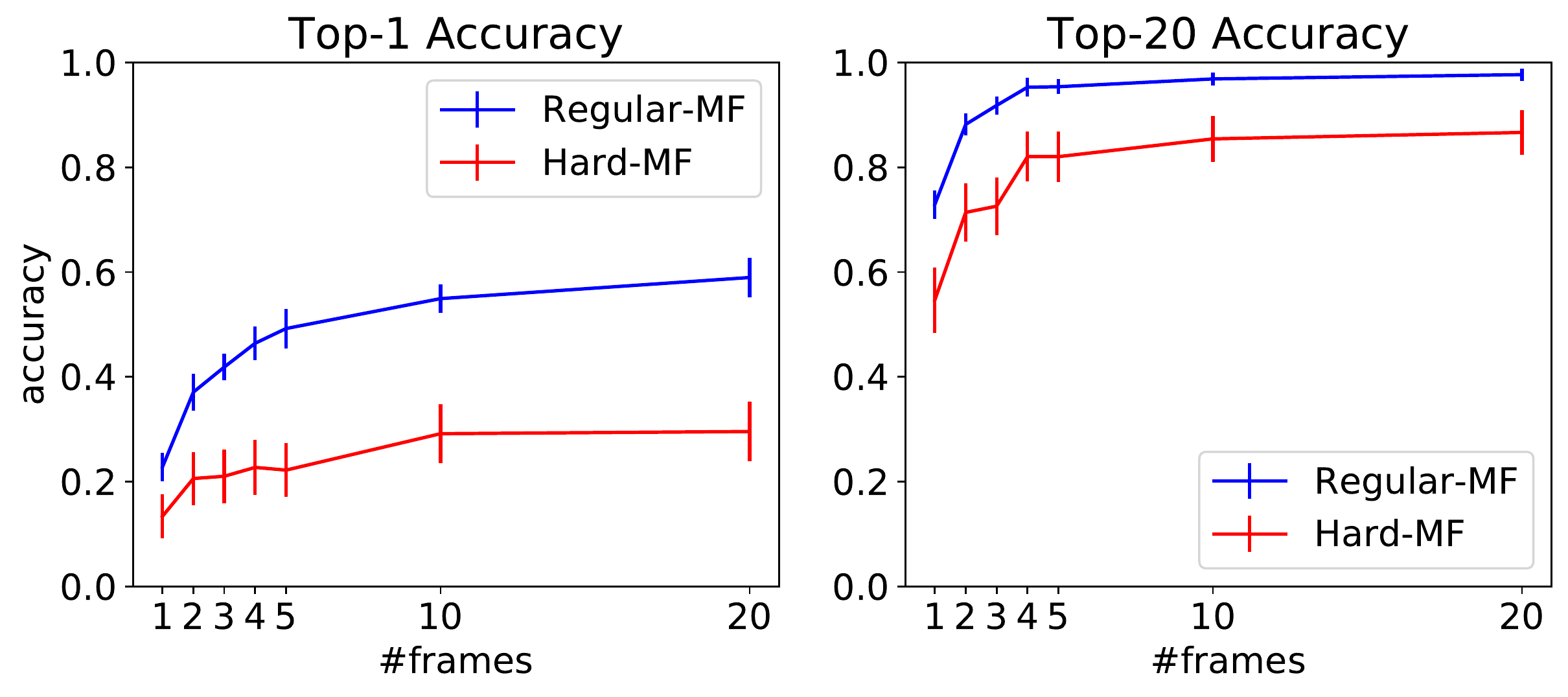}
            \caption{Plot of the \pname retrieval accuracy (y-axis) using different numbers of frames (x-axis) for aggregation.
            Error bars represent standard deviation of the accuracy. }
            \label{fig:ablation_nframes}
            
        \end{figure}

    Another experiment regards the length of the sequences. Fig.~\ref{fig:ablation_nframes} reports, with the associated error bars, the performance of \pname when increasing the number of frames from 1 to 20. As expected, the curves for both partitions, at both the top-1 and top-20 are increasing, with the ``Hard'' partition showing a plateau after 10 frames, while the ``Regular'' partition seem to benefit systematically. The reason could be that ``Hard'' sequences are dramatically noisy, and adding more frames will augment the clutter we need to deal with, while the ``Regular'' ones benefit because of the fine grained details which characterize the partition. Comparative performances when varying the sequence's length against other approaches are in Tab.~\ref{tab:frame_res}. Notably, Asymnet~\cite{cheng2017video2shop} does not reach our results \emph{even when doubling the number of input frames}.

        \begin{table}
        \centering
        \scriptsize
        \begin{tabular}{l| c c c c c}
         \toprule 
            \hfil\textbf{Categories}  & \textbf{NVAN} &\hfil\textbf{VKD} &\hfil\textbf{MGH} &\textbf{AsymNet}&  \hfil\textbf{SEAM}\\
           &  \hfil\cite{liu2019spatially} & \hfil\cite{porrello2020robust} &\hfil\cite{Yan_2020_CVPR} & \hfil\cite{cheng2017video2shop}  & \hfil\textbf{M-RCNN}\\
            \hline\hline
            \textit{Short Sl. Shirt} & \textbf{\hfil.46(.08)}& \hfil.20(.06)& \hfil.39(.06)&  \hfil.35(.07)&\hfil .43(.08)\\ \hline
            \textit{Long Sl. Shirt} & \hfil.38(.04)& \hfil.44(.04)& \hfil.41(.04)& \hfil.45(.04)& \hfil \textbf{.44(.04)}\\ \hline
            \textit{Short Sl. Outw.}  & \hfil.34(.20)& \hfil.23(.16)& \hfil.33(.18)& \hfil.35(.19)& \hfil\textbf{.42(.20)}\\ \hline
            \textit{Long Sl. Outw.} & \hfil.40(.10)& \hfil.43(.10)& \hfil.42(.10)& \hfil.36(.10)& \hfil \textbf{.46(.10)}\\ \hline
            \textit{Vest} & \hfil\textbf{.42(.12)}& \hfil.10(.07)& \hfil.24(.09)& \hfil.27(.11)& \hfil .31(.11)\\ \hline
            \textit{Sling} & \hfil.30(.19)& \hfil.16(.16)& \hfil.33(.18)& \hfil.32(.18) & \hfil\textbf{.36(.19)}\\ \hline
            \textit{Shorts} & \hfil.19(.13)& \hfil.27(.15)& \hfil.22(.13)& \hfil.25(.13) & \hfil\textbf{.39(.15)}\\ \hline
            \textit{Trousers} & \hfil.37(.05)& \hfil.28(.05)&  \hfil.35(.05)& \textbf{\hfil.45(.06)}& \hfil .39(.05)\\ \hline
            \textit{Skirt} & \hfil.40(.08)& \hfil.52(.08)& \hfil.47(.08)& \hfil.39(.08)&  \hfil\textbf{.56(.09)}\\ \hline
            \textit{Short Sl. Dress} & \hfil.34(.10)& \hfil.54(.11)&  \hfil.35(.10)&\hfil.45(.11)& \hfil \textbf{.73(.10)}\\ \hline
            \textit{Long Sl. Dress} & \hfil.37(.07)& \hfil.63(.07)&  \hfil.36(.07)&\hfil.57(.07)& \hfil \textbf{.68(.07)}\\ \hline
            \textit{Vest Dress} & \hfil.39(.09)& \hfil.49(.09)& \hfil.37(.09)& \hfil.42(.08) & \hfil\textbf{.64(.09)}\\ \hline
            \textit{Sling Dress} & \hfil.42(.14)& \hfil.39(.14)& \hfil.42(.13)& \hfil.32(.13)& \hfil \textbf{.69(.14)}\\ \hline
            \hline
             \textbf{All Classes}  &  \hfil.38(.02) & \hfil.40(.02) & \hfil.40(.02) & \hfil.42(.02)&  \hfil\textbf{.49(.02)}\\ \bottomrule
        \end{tabular}
        \vspace{0.5em}
        \caption{\label{tab:results_categories} Top-1 retrieval accuracy (and standard deviation) on MovingFashion for the 14 different item classes.    }
        \vspace{-0.5em}
        \end{table}

        \begin{table}[h]
            \centering
            \footnotesize
            \begin{tabular}{l|c c c}
            \hline
                \textbf{Method} & \textbf{5 Frames} & \textbf{10 Frames} & \textbf{20 Frames}  \\ \hline
                NVAN~\cite{liu2019spatially} & 0.35 & 0.38 & 0.39 \\ \hline
                VKD~\cite{porrello2020robust} & 0.36 &0.40 &0.43  \\ \hline
                MGH~\cite{Yan_2020_CVPR} & 0.36& 0.38& 0.40 \\ \hline
                AsymNet~\cite{cheng2017video2shop} & 0.37& 0.42& 0.44 \\ \hline
                \textbf{SEAM Match-RCNN} & \textbf{0.43} & \textbf{0.49}& \textbf{0.52} \\ \hline
            \end{tabular}
            \vspace{0.5em}
            \caption{\small\label{tab:frame_res}  Top-1 accuracy on MovingFashion, with different number of frames. }
            \vspace{-0.5cm}
                
        \end{table}
    \subsection{Experiments on unrelated sets of images}\label{sec:expmdf2}
    MovingFashion has street videos which depict clothing items in a variety of scenarios: indoor, outdoor, etc. We are interested in bringing this variety to the extreme, answering the following question: how does \pname behave when the street video sequence is formed by a few totally unrelated frames? 
    In order to perform these experiments, we build Multi-DeepFashion2 from DeepFashion2 using the pairings between shop images and street sequences composed of multiple corresponding street images. 
    The total pairings amount to 11K, each one composed of an image sequence (6 frames on average) sampled from different sources, along with the corresponding shop image. 
    Results are in Tab.~\ref{tab:results2}. Please note that, in order to be consistent with the 10-frames street sequence length we generate random repetitions for all the approaches given the smaller set of diverse images. The numbers indicate a decrease in general performance (less distinctive frames, more shop items); even in this case, we perform better than AsymNet. 
        
        \begin{table}
        \centering
        \scriptsize
        \begin{tabular}{m{88pt}| m{22pt} m{22pt} m{22pt} m{22pt} }
            \toprule
            \textbf{Method} & \hfil\textbf{T-1} &\hfil\textbf{T-5} & \hfil\textbf{T-10}  & \hfil\textbf{T-20}\\ \hline\hline
            Max-Confidence     &.19\var{.014}&  .44\var{.020} &.54\var{.020} &  .66\var{.017} \\ \hline
             Max Matching~\cite{cheng2017video2shop}    & .14\var{.015}      & .45\var{.020} & .61\var{.019}  &  .75\var{.017}\\ \hline
             \hline
             NVAN (2019)~\cite{liu2019spatially}&       .22\var{.019}    &  .37\var{.019} &    .43\var{.018} &    .49\var{.019}\\ \hline
            VKD (2020)~\cite{porrello2020robust}&       .21\var{.014}    & .27\var{.017} &    .33\var{.017} &    .38\var{.018}\\ \hline
            MGH (2020)~\cite{Yan_2020_CVPR}&       .22\var{.016}    & .34\var{.018} &    .39\var{.017} &    .45\var{.019}\\ \hline
            \hline

            AsymNet [GT]~\cite{cheng2017video2shop}&       .21\var{.016}    &  .50\var{.019} &    .62\var{.017} &    .74\var{.017}\\ \hline
            AsymNet (2017)~\cite{cheng2017video2shop}&       .18\var{.018}    &  .43\var{.020} &    .57\var{.020} &    .70\var{.018}\\ \hline
            AsymNet [AVG]&       .16\var{.016}    &  .41\var{.020} &    .54\var{.020} &    .68\var{.018}\\ \hline
            AsymNet [MAX]&       .15\var{.017}    &  .42\var{.020} &    .56\var{.019} &    .70\var{.017}\\ \hline\hline
            Average Distance~\cite{cheng2017video2shop}    & .22\var{.017}      &  .49\var{.020} &  .63\var{.019}  &  .74\var{.018}\\ \hline
            \textbf{\pname w/o NLB, $g$}    & .20\var{.016}      &  .47\var{.021} &  .60\var{.021}  &  .71\var{.019}\\ \hline
             \textbf{\pname [GT]}    & .30\var{.013}      &  .58\var{.016} &  .67\var{.016}  & .76\var{.014} \\ \hline
            \textbf{\pname}    & \textbf{.28\var{.018}}      &  \textbf{.54\var{.020}} &  \textbf{.66\var{.019}}  & \textbf{.76\var{.017}} \\ \bottomrule
        \end{tabular}
        \vspace{0.5em}
        \caption{\label{tab:results2} Video-to-Shop retrieval results on MultiDeepFashion2.
        Note: T-K means Top-K Accuracy. }
        \vspace{-0.5em}
        \end{table}

    \subsection{Experiments on the attention mechanism}\label{sec:expatt}
\begin{figure}[t!]
            \centering
            \includegraphics[width=\linewidth]{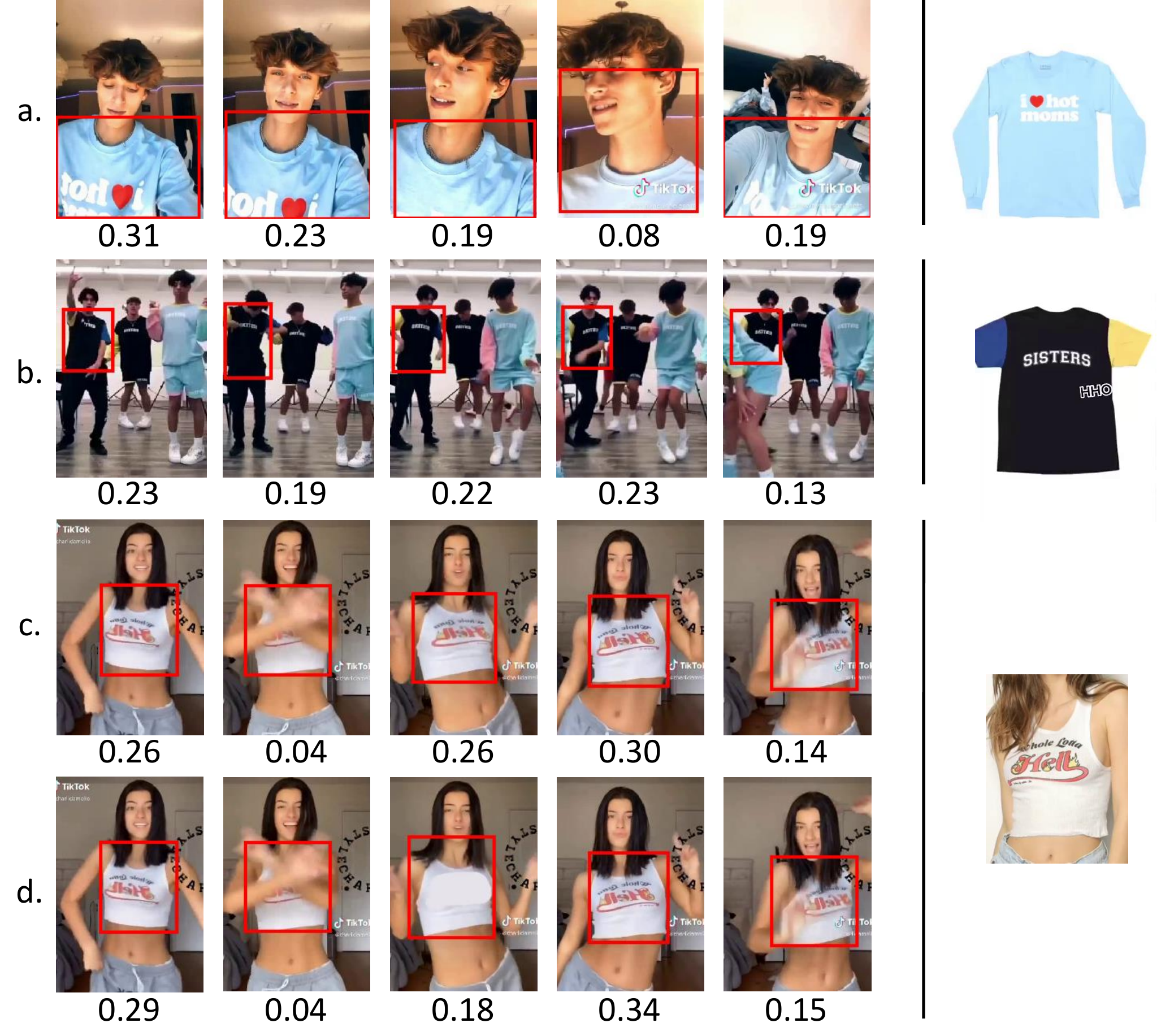}
            \caption{Qualitative observations on the attention behaviour.
            On the left, for each video sequence we show the detection bounding boxes and the computed attention score. On the right the paired shop item.
            }
            \label{fig:ablation_attention}
        \end{figure}

        The ablation studies of Table~\ref{tab:results} clearly show that the attention layers play a crucial role for the \pname performance. Here we explain their role qualitatively and quantitatively.          
        In Fig.~\ref{fig:ablation_attention} we report the attention values obtained after the application of the attention layer $g$ to the output of the self-attention layer $NLB$ of Sec.~\ref{sec:featuresextraction}, i.e., $g(NLB(x))$. On row a), one can note that the attention is high when the heart logo is visible (0.31, 0.23 in the first two frames) and it goes down when it vanishes, despite the light blue shirt (last frame) being very similar area-wise. This means that the mechanism considers the heart logo as important for retrieval. On the second row b), the effect of an occlusion in the attention score (last frame). On the third row c), a white top with a logo gives a stable attention score (around 0.28). We manually cover the logo in the third frame, causing a clear decrease in the attention, uniformly increasing the ones highlighting the logo.

\begin{figure}[t!]
            \centering
            \includegraphics[width=\linewidth]{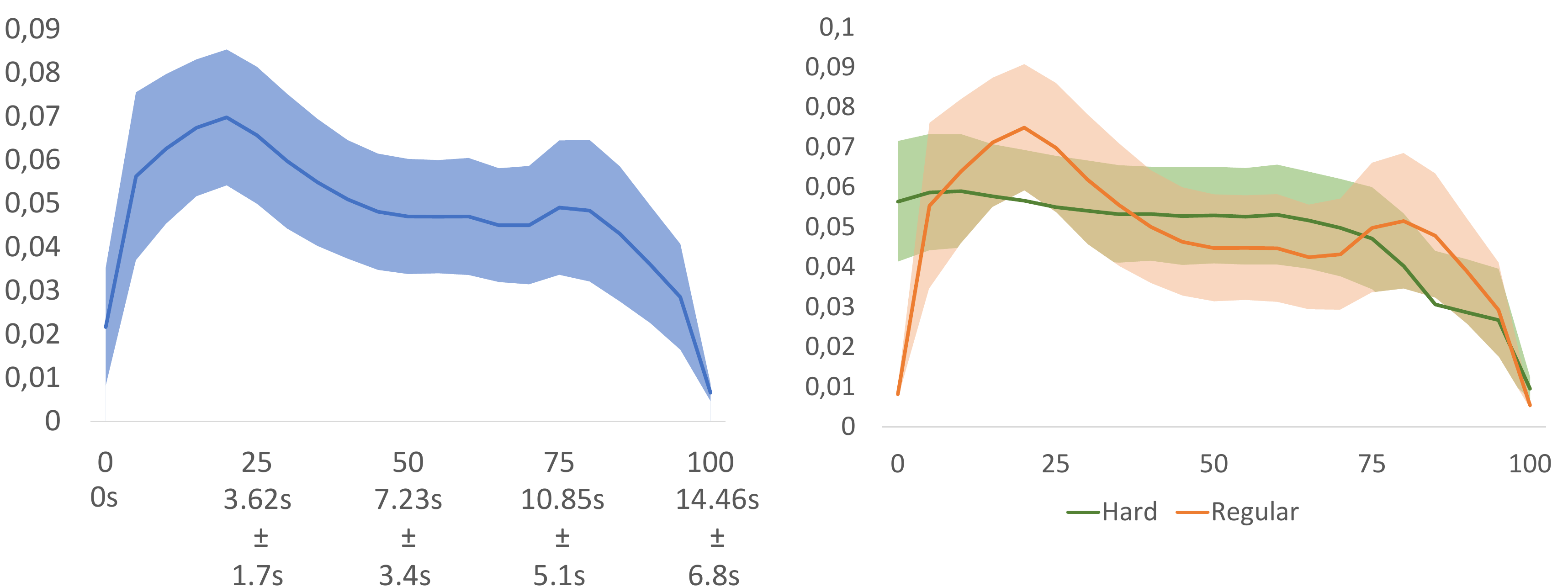}
            \caption{Mean attention score every 5 percentiles of the video length. For each video we sampled 21 equally spaced frames. On the left we report the average attention (y-axis) and frame-timing information (x-axis labels) for the whole MovingFashion dataset. On the right for the Regular and Hard subsets. We show error bands for the standard deviation. 
            }
            \label{fig:ablation_prc_attention}
            \vspace{-0.5cm}
        \end{figure}

        Finally, driven by best practices in social video editing~\cite{bestpractices2019}, which state that a video message has to deliver its main content in the first 6 seconds to trigger the observers' attention, we calculate the attention every 5 percentiles on all the Moving fashion sequences, producing the curves in Fig.~\ref{fig:ablation_prc_attention}a) (on the whole MovingFashion dataset) and on the separate partitions Fig.~\ref{fig:ablation_prc_attention}b. Surprisingly, the data confirms this rule, showing a clear (Fig.~\ref{fig:ablation_prc_attention}a) peak around the first quartile (definitely within 6 seconds), then a decrease and a later increase with a local maximum on the fourth quartile. The same holds for the two separate partitions (Fig.~\ref{fig:ablation_prc_attention}b)), with less emphasis on the ``Hard partition''. The reason lies in the nature of the Net-A-Porter videos, which in many cases show the entire clothing item in the beginning of the sequence, with the model that moves subsequently, zooming up to critical detail (the belt for the shorts) towards the end (second peak). On the ``Hard'' partition, the attention for the clothing items is higher in the beginning, since the actors present their outfit and then exhibit their performance (dancing, gymnastics etc.), concluding in both the cases with uninteresting details clothing wise. 
           
    \section{Conclusions}
        Our \pnamenospace, trained on the new MovingFashion dataset, provides a strong baseline that shows video-to-shop matching can be performed on videos in the wild like TikToks, possibly unveiling fashion trends directly from social platforms and consequently attracting big fashion players. 
    \section*{Acknowledgments}
This work is partially supported by the Italian MIUR through PRIN 2017 - Project Grant 20172BH297: I-MALL - improving the customer experience in stores by intelligent computer vision, and by the project of the Italian
Ministry of Education, Universities and Research (MIUR) "Dipartimenti di Eccellenza 2018-2022". Thanks also to Giovanni Masotto for the collaboration in creating MovingFashion.

{\small
\bibliographystyle{ieee_fullname}
\bibliography{egbib}

\begin{thebibliography}{10}\itemsep=-1pt

\bibitem{cheng2017video2shop}
Zhi-Qi Cheng, Xiao Wu, Yang Liu, and Xian-Sheng Hua.
\newblock Video2shop: Exact matching clothes in videos to online shopping
  images.
\newblock In {\em Proceedings of the IEEE Conference on Computer Vision and
  Pattern Recognition}, pages 4048--4056, 2017.

\bibitem{duffett2020youtube}
Rodney Duffett.
\newblock The youtube marketing communication effect on cognitive, affective
  and behavioural attitudes among generation z consumers.
\newblock {\em Sustainability}, 12(12):5075, 2020.

\bibitem{gao2018revisiting}
Jiyang Gao and Ram Nevatia.
\newblock Revisiting temporal modeling for video-based person reid.
\newblock {\em arXiv preprint arXiv:1805.02104}, 2018.

\bibitem{ge2019deepfashion2}
Yuying Ge, Ruimao Zhang, Xiaogang Wang, Xiaoou Tang, and Ping Luo.
\newblock Deepfashion2: A versatile benchmark for detection, pose estimation,
  segmentation and re-identification of clothing images.
\newblock In {\em Proceedings of the IEEE conference on computer vision and
  pattern recognition}, pages 5337--5345, 2019.

\bibitem{bestpractices2019}
Maxwell Golling.
\newblock Facebook video ads: Best practices for 2019, 2018.

\bibitem{hadi2015buy}
M Hadi~Kiapour, Xufeng Han, Svetlana Lazebnik, Alexander~C Berg, and Tamara~L
  Berg.
\newblock Where to buy it: Matching street clothing photos in online shops.
\newblock In {\em Proceedings of the IEEE international conference on computer
  vision}, pages 3343--3351, 2015.

\bibitem{li2018diversity}
Shuang Li, Slawomir Bak, Peter Carr, and Xiaogang Wang.
\newblock Diversity regularized spatiotemporal attention for video-based person
  re-identification.
\newblock In {\em Proceedings of the IEEE Conference on Computer Vision and
  Pattern Recognition}, pages 369--378, 2018.

\bibitem{liu2019spatially}
Chih-Ting Liu, Chih-Wei Wu, Yu-Chiang~Frank Wang, and Shao-Yi Chien.
\newblock Spatially and temporally efficient non-local attention network for
  video-based person re-identification.
\newblock In {\em British Machine Vision Conference}, 2019.

\bibitem{liu2016deepfashion}
Ziwei Liu, Ping Luo, Shi Qiu, Xiaogang Wang, and Xiaoou Tang.
\newblock Deepfashion: Powering robust clothes recognition and retrieval with
  rich annotations.
\newblock In {\em Proceedings of the IEEE conference on computer vision and
  pattern recognition}, pages 1096--1104, 2016.

\bibitem{porrello2020robust}
Angelo Porrello, Luca Bergamini, and Simone Calderara.
\newblock Robust re-identification by multiple views knowledge distillation.
\newblock In {\em The European Conference on Computer Vision (ECCV)}, 2020.

\bibitem{shen2019person}
Yantao Shen, Tong Xiao, Shuai Yi, Dapeng Chen, Xiaogang Wang, and Hongsheng Li.
\newblock Person re-identification with deep kronecker-product matching and
  group-shuffling random walk.
\newblock {\em IEEE transactions on pattern analysis and machine intelligence},
  2019.

\bibitem{wang2018non}
Xiaolong Wang, Ross Girshick, Abhinav Gupta, and Kaiming He.
\newblock Non-local neural networks.
\newblock In {\em Proceedings of the IEEE conference on computer vision and
  pattern recognition}, pages 7794--7803, 2018.

\bibitem{xuan2020improved}
Hong Xuan, Abby Stylianou, and Robert Pless.
\newblock Improved embeddings with easy positive triplet mining.
\newblock In {\em The IEEE Winter Conference on Applications of Computer
  Vision}, pages 2474--2482, 2020.

\bibitem{Yan_2020_CVPR}
Yichao Yan, Jie Qin, Jiaxin Chen, Li Liu, Fan Zhu, Ying Tai, and Ling Shao.
\newblock Learning multi-granular hypergraphs for video-based person
  re-identification.
\newblock In {\em Proceedings of the IEEE/CVF Conference on Computer Vision and
  Pattern Recognition (CVPR)}, June 2020.

\bibitem{zhaodress}
Hongrui Zhao, Jin Yu, Yanan Li, Donghui Wang, Jie Liu, Hongxia Yang, and Fei
  Wu.
\newblock Dress like an internet celebrity: Fashion retrieval in videos.
\newblock In {\em proceedings of the International Joint Conferences on
  Artificial Intelligence}, pages 1054--1060, 07 2020.

\end{thebibliography}
}
\clearpage
\pagebreak
\appendix

       \begin{table*}[t!]
        \centering
        \begin{tabular}{m{120pt}|l l l l | l l l l| l l l l}
            \toprule
            \textbf{Method} & \multicolumn{4}{l|}{\hfil\textbf{MovingFashion}} & \multicolumn{4}{l|}{\hfil \textbf{Regular-MovingFashion}}& \multicolumn{4}{l}{\hfil \textbf{Hard-MovingFashion}}\\
             & \hfil\textbf{T-1} & \hfil\textbf{T-5} & \hfil\textbf{T-10} & \hfil\textbf{T-20} & \hfil\textbf{T-1} & \hfil\textbf{T-5} & \hfil\textbf{T-10} & \hfil\textbf{T-20} & \hfil\textbf{T-1} & \hfil\textbf{T-5} & \hfil\textbf{T-10} & \hfil\textbf{T-20}\\ \hline

             SFM-First     & 0.20      &  0.43 &  0.52&  0.63 & 0.21 & 0.44 &  0.53  &  0.64 & 0.16 &  0.41 &  0.52  & 0.62\\ \hline
             SFM-1qrt     &0.25      & 0.53 & 0.66& 0.77 & 0.29  &  0.58 &  0.71  &  0.82 & 0.15      &  0.37 &  0.51  &  0.63\\ \hline
             SFM-Median     &0.23     & 0.48 & 0.61& 0.75 & 0.26      &  0.53 &  0.66  &  0.79 & 0.17      &  0.33 &  0.47  &  0.65\\ \hline
             SFM-3qrt     &0.21     & 0.47 & 0.60& 0.72 & 0.24      &  0.53 &  0.66  &  0.77& 0.13      &  0.29 &  0.42  &  0.57\\ \hline
             SFM-Last     &0.11    & 0.31 & 0.41& 0.53 & 0.14  &  0.35 &  0.46  &  0.58 & 0.05      &  0.18 &  0.27  &  0.36\\ \hline\hline
             
             EPHN-First (2020)~\cite{xuan2020improved} &0.15&	0.34&	0.44&	0.53&	0.16&	0.36&	0.46&	0.55&	0.11&	0.27&	0.37&	0.47\\ \hline
             EPHN-1qrt & 0.24&	0.45&	0.55&	0.65&	0.28&	0.51&	0.62&	0.72&	0.13&	0.24&	0.32&	0.42 \\ \hline
             EPHN-Median &0.27&	0.49&	0.58&	0.66&	0.32&	0.57&	0.67&	0.74	&0.10&	0.24&	0.32&	0.42 \\ \hline
             EPHN-3qrt &0.24&	0.47&	0.55&	0.65&	0.29&	0.55&	0.64&	0.74&	0.09&	0.21&	0.29&	0.40 \\ \hline
             EPHN-Last &0.17&	0.33&	0.41&	0.49&	0.20&	0.39&	0.47&	0.56&	0.07&	0.15&	0.19&	0.27 \\\hline\hline
             
             KPM-First (2019)~\cite{shen2019person}& 0.19 & 0.40 & 0.51 & 0.61 & 0.22 & 0.45 & 0.56 & 0.67 & 0.09 & 0.26 & 0.33 & 0.45\\ \hline
             KPM-1qrt & 0.27 & 0.48 & 0.60 & 0.71 & 0.32 & 0.56 & 0.69 & 0.80 & 0.12 & 0.24 & 0.33 & 0.45 \\ \hline
             KPM-Median & 0.24 & 0.48 & 0.59 & 0.69 & 0.27 & 0.55 & 0.67 & 0.78 & 0.12 & 0.25 & 0.35 & 0.43 \\ \hline
             KPM-3qrt & 0.23 & 0.46 & 0.56 & 0.69 & 0.27 & 0.53 & 0.65 & 0.76 & 0.10 & 0.22 & 0.28 & 0.39 \\ \hline
             KPM-Last & 0.16 & 0.35 & 0.45 & 0.55 & 0.20 & 0.41 & 0.53 & 0.65 & 0.05 & 0.14 & 0.19 & 0.23 \\\hline\hline
             
             \textbf{SEAM Match-RCNN}    & \textbf{0.49}      &  \textbf{0.80} &  \textbf{0.89} &  \textbf{0.94} & \textbf{0.55}      &  \textbf{0.86} &  \textbf{0.94}  &  \textbf{0.97}& \textbf{0.30}      &  \textbf{0.62} &  \textbf{0.76}  &  \textbf{0.87} \\ \bottomrule

        \end{tabular}
        \vspace{0.5em}
        \caption{\label{tab:results_mf_single} SEAM Match-RCNN retrieval results on MovingFashion compared with Single-frame approaches.
        Note: T-K means Top-K Accuracy.}
        \vspace{-1em}
        \end{table*}
        \begin{table*}[t!]
        \centering
        \begin{tabular}{m{120pt}|l l l l | l l l l| l l l l}
            \toprule
            \textbf{Method} & \multicolumn{4}{l|}{\hfil\textbf{MovingFashion}} & \multicolumn{4}{l|}{\hfil \textbf{Regular-MovingFashion}}& \multicolumn{4}{l}{\hfil \textbf{Hard-MovingFashion}}\\
             & \hfil\textbf{T-1} & \hfil\textbf{T-5} & \hfil\textbf{T-10} & \hfil\textbf{T-20} & \hfil\textbf{T-1} & \hfil\textbf{T-5} & \hfil\textbf{T-10} & \hfil\textbf{T-20} & \hfil\textbf{T-1} & \hfil\textbf{T-5} & \hfil\textbf{T-10} & \hfil\textbf{T-20}\\ \hline

             Max Confidence     & 0.29      &  0.59 &  0.72 &  0.83 & 0.31       &  0.63  &  0.76 &  0.86 & 0.21      &  0.46 &  0.60  &  0.71\\ \hline
             Max Matching    & 0.26     &  0.60 &  0.74  &  0.85 & 0.29      &  0.65 &  0.79  &  0.89 & 0.17      &  0.44 &  0.58  &  0.74\\ \hline
             Average Match-RCNN~\cite{cheng2017video2shop}   & 0.39      &  0.73 &  0.84  &  0.91 &  0.43      &  0.79 &  0.88  &  0.94 & 0.24      &  0.56 &  0.70  &  0.81\\ \hline
            Average Descriptor  & 0.37      &  0.72 &  0.86  &  0.93 & 0.42 &  0.78 &  0.90  &  0.95 & 0.21 &  0.57 &  0.75  &  0.85 \\
             \hline \hline
             
             EPHN-MaxConf (2020)~\cite{xuan2020improved}& 0.22&	0.43&	0.55&	0.65&	0.26&	0.50&	0.61	&0.71&	0.10&	0.22&	0.34&	0.44 \\ \hline
             EPHN-MaxMatching & 0.35 &	0.59&	0.67	&0.74&	0.42&	0.68&	0.76&	0.81&	0.14&	0.32&	0.41&	0.52 \\ \hline
             EPHN-AvgMatching &0.31 &	0.55	&0.64&	0.73&	0.37&	0.64&	0.73&	0.81&	0.11&	0.28&	0.37&	0.48 \\ \hline
             EPHN-AvgDescriptor & 0.22&	0.43&	0.52&	0.61&	0.26&	0.49	&0.58&	0.68&	0.10&	0.24&	0.33& 0.43 \\ \hline\hline
             
             KPM-MaxConf (2019)~\cite{shen2019person} & 0.25 & 0.47 & 0.57 & 0.68 & 0.30 & 0.54 & 0.65 & 0.77 & 0.11 & 0.25 & 0.32 & 0.43  \\ \hline
             KPM-MaxMatching & 0.30 & 0.54 & 0.66 & 0.75 &  0.36 & 0.61 & 0.73 & 0.82 & 0.13 & 0.32 & 0.42 & 0.53 \\ \hline
             KPM-AvgMatching & 0.34 & 0.58 & 0.68 & 0.77 & 0.40 & 0.68 & 0.78 & 0.86 & 0.15& 0.28 & 0.38 & 0.48 \\ \hline
             KPM-AvgDescriptor & 0.34 & 0.58 & 0.69 & 0.77 & 0.40 & 0.68 & 0.78 & 0.86 & 0.15 & 0.28 & 0.38 & 0.48 \\ \hline\hline

             \textbf{SEAM Match-RCNN}    & \textbf{0.49}      &  \textbf{0.80} &  \textbf{0.89} &  \textbf{0.94} & \textbf{0.55}      &  \textbf{0.86} &  \textbf{0.94}  &  \textbf{0.97}& \textbf{0.30}      &  \textbf{0.62} &  \textbf{0.76}  &  \textbf{0.87}\\
            
            \bottomrule
             
               \end{tabular}
        \vspace{0.5em}
        \caption{\label{tab:results_mf_multi} SEAM Match-RCNN retrieval results on MovingFashion compared with Multi-frame approaches.
        Note: T-K means Top-K Accuracy.}
        \vspace{-1em}
    \end{table*}

         \begin{table}[t!]
        \centering
        \small
        \begin{tabular}{m{110pt}|l l l l}
            \toprule
            \textbf{Method} & \multicolumn{4}{l}{\hfil\textbf{MultiDeepFashion2}}\\
             & \hfil\textbf{T-1} & \hfil\textbf{T-5} & \hfil\textbf{T-10} & \hfil\textbf{T-20}\\ \hline
             Max Confidence & 0.19 &  0.44 &  0.54 &  0.66\\ \hline
             Max Matching    & 0.14     &  0.45 &  0.61  &  0.75\\ \hline
             Average Match-RCNN~\cite{cheng2017video2shop}   & 0.22      &  0.49 &  0.63  &  0.74\\ \hline
             Average Descriptor &0.20      &  0.48 &  0.60  &  0.71 \\
             \hline \hline
             EPHN-MaxConf (2020)~\cite{xuan2020improved}       & 0.11 &	0.19 &	0.24 &	0.29\\ \hline
             EPHN-MaxMatching   & 0.11 &	0.21 &	0.26 &	0.33 \\ \hline
             EPHN-AvgMatching   & 0.16 &	0.29 &	0.34 &	0.41\\ \hline
             EPHN-AvgDescriptor & 0.12 &	0.22 &	0.27 &	0.33\\ 
             \hline\hline
             KPM-MaxConf (2019)~\cite{shen2019person}    &  0.09 & 0.20 & 0.25 & 0.30 \\\hline
             KPM-MaxMatching &  0.08 & 0.16 & 0.21 & 0.28 \\\hline
             KPM-AvgMatching  & 0.10 & 0.20 & 0.25 & 0.32 \\ \hline
             KPM-AvgDescriptor & 0.13 & 0.25 & 0.33 & 0.40 \\ \hline\hline
             \textbf{SEAM Match-RCNN}    & \textbf{0.28}      &  \textbf{0.54} &  \textbf{0.66}  & \textbf{0.76} \\ \bottomrule

        \end{tabular}
        \vspace{0.5em}
        \caption{\label{tab:results_df2_multi} SEAM Match-RCNN retrieval results on MultiDeepFashion2 compared with Multi-frame approaches.
        Note: T-K means Top-K Accuracy.}
        \vspace{-1em}
        \end{table}

\section{MovingFashion Additional Details}\label{sec:data}
\subsection{Image and video collection}    
    In this section we give further details on the process of data collection and annotation of Moving Fashion. 

    Regarding the data collected from the Net-A-Porter website, the data labeling was a long, yet linear process, the only issue being the removal of classes not in the DeepFashion2 taxonomy, in particular \emph{shoes} (deserving of a specific fashion taxonomy) and \emph{jewelry} (due to the lack of a shared and widely accepted aesthetical taxonomy). For the remaining classes, the association to the specific DF2 taxonomy was direct. 
    
    Plenty more work was required for the data downloaded from Instagram. In order to to download the data, the Instaloader\footnote{\url{https://instaloader.github.io/}} tool was employed. We manually selected a list of hashtags and profiles with a lot of content, i.e. a lot of videos paired with fashion products for sale. Through the use of the tool, we downloaded posts containing videos only based on the previously mentioned hashtags and profiles. The layout of these videos was standard for the vast majority of them: the frame was divided vertically in two parts, one with just a still picture of the shop product and one with the video itself.
    
    We manually annotated these videos by following these steps:
    \begin{itemize}
        \item We checked that the product actually appears in the video, since in some cases the item never appears or appears very briefly in the frame; sometimes the item is in a different color than the one in the shop image.
        \item We drew a bounding box around the area of the shop item(s), taking care to include as few other items as possible. 
        \item We drew another bounding box around the area of the video.
    \end{itemize}
    Using these annotations we crop the street videos and shop images. This results in pairings, where in some cases we have more than one shop item associated with a street video.
    
    Next, we dealt with duplicates of shop products. In some cases the same product is showcased in multiple videos by different users, but fortunately, the shop image used in such videos is the same. We leveraged this fact to perform a duplicate search for all the shop images. Products that were found to be duplicates were merged, creating pairings where for one shop product multiple videos are associated. To perform this search, for each product we searched for duplicates using a pre-existing tool\footnote{\url{https://github.com/umbertogriffo/fast-near-duplicate-image-search}} that employs Perceptual Hash. However we found out that in order to have a very high recall, this process also includes a lot of false positives. To perform a more thorough search, we tried an Image Registration technique using the RANSAC algorithm between each shop image and the duplicate candidates found using the tool. We tried to estimate a Similarity Transform, to account for translations and scaling (as is the case for these images). We then put a threshold on average pixel difference to separate between duplicates and non duplicates. Since no Python libraries that implement RANSAC are available, it was performed using a custom script.
    
    To make sure that MovingFashion respects the privacy of social media users, we have rendered any face in the videos blurred using a publicly available, face blurring tool\footnote{\url{https://github.com/ORB-HD/deface}}.
        
\subsection{Tracklet generation}     
    As described in the paper, for all data, noisy tracklet annotations are available.
    In order to create them:
    \begin{itemize}
        \item Our SEAM Match-RCNN is trained on the data \emph{using only video-image pairing annotations}. This results in a model where the Single-frame Matching Head can be effectively used for precisely tracking each item.
        \item We use the trained model to build a set of tracklets for each video.
        \item We manually go over each video and select the tracklets that contain the paired shop item, merging them if they are disjointed (this happens when an item is occluded completely or disappears from the frame and two separate tracklets are built).
    \end{itemize}
    
    The resulting tracklets are then saved. While for our approach, no tracklet annotations are used during training, they are used for all the comparative approaches. They are considered as equivalent to ours (the detector and the tracker are the same). It can be argued that they are actually better than ours as they are produced after the last epoch of training, while for our approach we start with a tracker that has not been trained yet. For the Person Re-ID approaches, the annotations are used to crop out part of the image according to the extracted bounding box. For detection based approaches, the bounding boxes are used as ground truth bounding boxes. The testing tracklets are used by all approaches for evaluation. During the SEAM Match-RCNN evaluation, they are used to select the tracklet among the ones produced automatically by the \emph{tracking procedure}.

\section{SEAM Match-RCNN computational complexity}\label{sec:complexity}
    In this Section we discuss the computational complexity of our proposed SEAM Match-RCNN. In particular we focus on the difference between the Single-frame Matching Head and the Multi-frame Matching Head.
    
    \subsection{Single-frame Matching Head}
    
        Let $TF$ be the time taken for computing features by using the $f$ function and $TM$ the time taken for computing matching between two feature vectors using $m$.
        
        Given a street image and a shop image, the cost of computing a matching between them, assuming that the detection from the street image has already been chosen in some way (for example by comparing it with a ground truth bounding box) is $2\times TF + TM$ (features computed for both street and shop are compared).
    
    \subsection{Multi-frame Matching Head}
        
        When extending to Multi-frame matching, the cost of tracking has to be taken into consideration. Obviously the time taken for feature computation increases linearly with the number of frames sampled from the video.
        
        As $\tilde{f}$ and $\tilde{m}$ are structured in the same way as $f$ and $m$, we can assert that $TF$ and $TM$ also apply to them.
        Given a street video sequence from which we sample $T$ frames, the cost of building all the possible tracklets (using the \emph{tracking procedure}, Sec~4-1 of the main paper) is related to the number of detections in each frame $K$ (to simplify notation we assume that there are exactly $K$ detections in each frame). First of all, Single-frame Matching Head features are computed, the time cost is $TF \times K \times T$.
        
        As a reminder, the tracking procedure consists of iteratively repeating the choice of \emph{pivot} and \emph{propagation}. The choice of the pivot is performed by choosing the most confident detection, so its cost is negligible as it is already included in the detection. The propagation consists of doing comparisons between the pivot features and all of the detection features in a frame. For a Single-frame the time necessary for the propagation step is $TM \times K$ (a matching for each detection). This procedure is repeated for all frames resulting in $TM \times K \times T$. This results in a single tracklet, that is excluded from the set of detections for the following iterations. As the iteration is repeated until there are no more detections, we can assume that repeating the propagation $K$ times results in a final cost of $TM \times K^2 \times T$. For the whole tracking procedure, the total time is $(TF \times K \times T) + (TM \times K^2 \times T)$.
    
        After tracklets are built, we can assume that the correct tracklet is chosen, for example by using the Intersection over Union with the ground truth tracklets (analogous to selecting the correct bounding box in the Match-RCNN). Given a sequence of detection of length $T$ (length of the video sequence), the cost of computing Multi-frame Matching features is again $TF \times T$. Then self-attetion with the Non-Local Block is performed, resulting in a time cost of $T^2 \times TSA$ ($TSA$ is the cost of computing self-attention between a pair of element in the sequence, usually a simple operation like a dot product). The attention score is then computed for each frame, with a cost of $T \times TA$ ($TA$ is the cost of computing the attention score, in our case a simple linear layer). Finally a weighted average pooling is performed and matching is computed between the aggregated descriptor and the shop feature vector ($TF + TM$). The final cost for aggregation is $(TF \times T) + (T^2 \times TSA) + (T \times TA) + (TF + TM)$.
        
    \subsection{Discussion}
        
        It is expected that the extension from Single-frame to Multi-frame will come with an increased cost, in relation to the number of frames. The tracking procedure is a necessary step for any possible Multi-frame approach, as detections from each frame need to be grouped in some way. The matching component increases quadratically with the maximum number of detections in each frame and linearly with the number of frames sampled from the sequence.
        
        The aggregation has a term that increases quadratically with the number of frames. For both of these, we have to take into consideration that we usually work with 10 samples and there are rarely many different people and clothing items in a video, so even with a quadratic complexity, the total effective time is relatively small. In our experiments, we never go over 2 seconds for the whole procedure, with the majority of the videos taking about 1 second to process.

\section{Additional experiments}\label{sec:add_results}
In Table~\ref{tab:results_mf_single}, we show the results of Single-frame baselines built on top of the Match-RCNN (the main building block of our SEAM Match-RCNN). In particular, SFM-1qrt  uses the frame at the first quartile of all the available frames of that sequence, SFM-median uses the median frame and so on. SFM stands for Single-frame match and is a short term for Match-RCNN.

The correspondent baselines are shown, adopting the Deep Kronecker-Product Matching (KPM)~\cite{shen2019person} and the Easy Positive Triplet Mining approach (EPHN)~\cite{xuan2020improved}. The rationale of this choice was to focus on Single-frame Re-Identification approaches and compare them to the Match-RCNN. This was done to enlarge the spectrum of possible comparative approaches, which have open-source code available. The idea of considering Re-ID approaches against street-to-shop techniques was also presented in the DPRNet paper~\cite{zhaodress}.

The inferiority of these baselines with respect of the Multi-frame of Table 2 in the main paper, and in particular with SEAM Match-RCNN, is evident and fully understandable. 

Notably, in almost all of the MovingFashion partitions (apart the regular one with EPHN), the $\cdot$-1qrt baseline gives the higher results, which seems to be in accord with the best practices in social media video editing, that is, that videos have to deliver their main message within approximately 6 seconds~\cite{bestpractices2019}.

As additional Multi-frame approaches, Table~\ref{tab:results_mf_multi} shows Max Confidence, Max Matching and Average Matching scores when considering the KPM~\cite{shen2019person} and the EPHN~\cite{xuan2020improved} as Single-frame method ingredients, in the same way that Match-RCNN was used to calculate Max Confidence, Max Matching and Average Matching from Table 2 of the main paper. 

Even in this case, SEAM Match-RCNN gives the best performance, showing an overall superiority of Match-RCNN as a Single-frame tool to aggregate visual clothing information. 

The same applies when it comes to MultiDeepFashion2 where we investigate only Multi-frame policies (Max Confidence, Matching, Avg Matching and Descriptor), since Single-frame policies do not have much sense, as the Single-frames are not part of a single sequence. Even in this case, SEAM Match-RCNN is the best alternative (Table~\ref{tab:results_df2_multi}).

\begin{figure*}[t!]
        \centering
        \includegraphics[width=\linewidth]{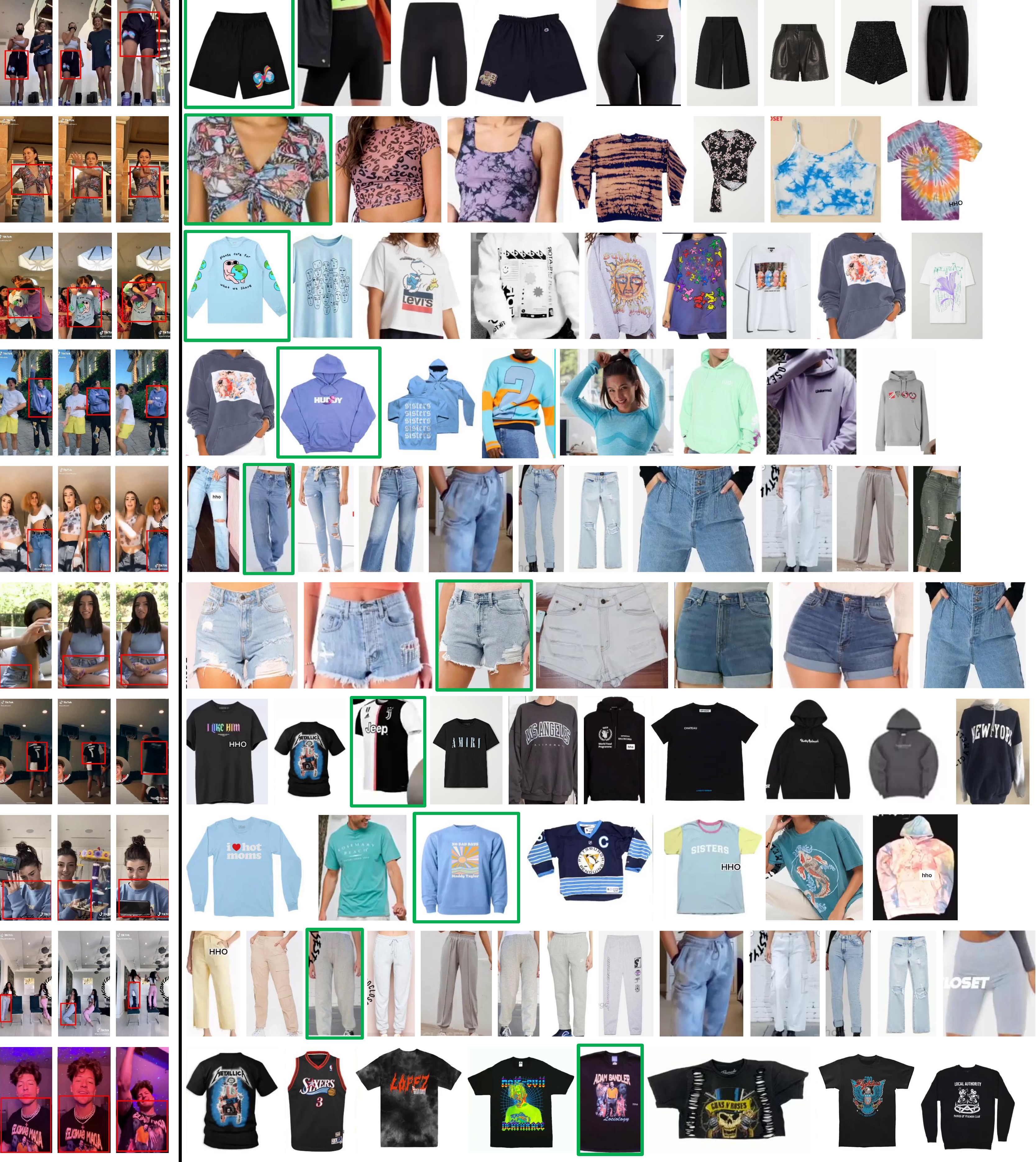}
        \caption{Qualitative retrieval results of SEAM Match-RCNN for the Hard-MovingFashion dataset. On the left, we show 3 frames sampled from the 10 frames used for aggregation. On the right the shop images retrieved starting from the closest match (left). The correct matches are represented with a green border.
        }
        \label{fig:retrievals1}
        \vspace{-1em}
    \end{figure*}

As additional \emph{qualitative} results, on Fig.~\ref{fig:retrievals1} results of SEAM Match-RCNN for the Hard-MovingFashion dataset are shown. Two types of considerations can be drawn: the first one is the variability of the videos, which here can be appreciated with more examples. Second, the retrieval results on the right display that SEAM Match-RCNN is capable of finding similar images, among a shop gallery that in some cases contains highly similar items (see for example the light gray trousers).

\begin{figure*}[t!]
        \centering
        \includegraphics[width=\linewidth]{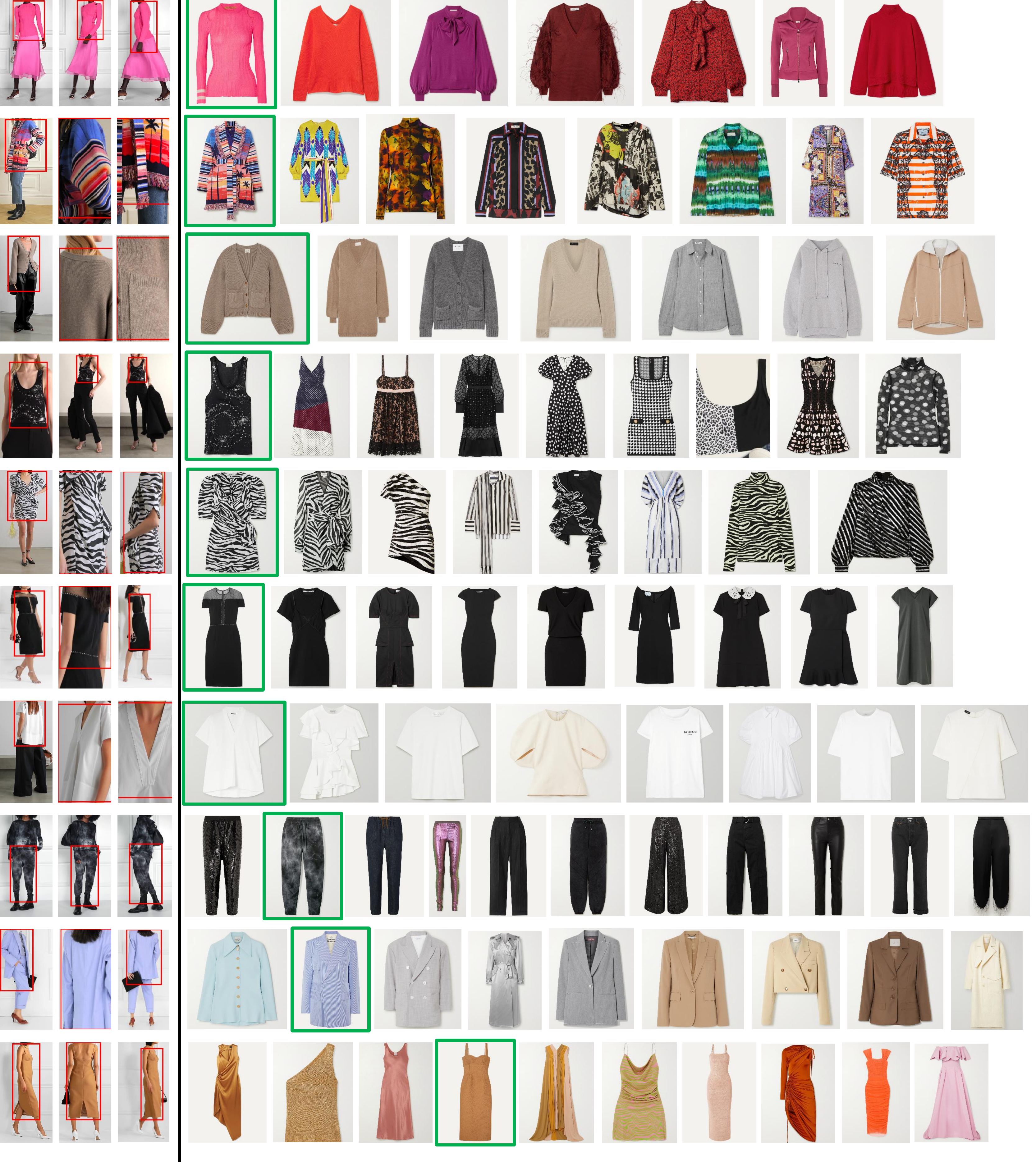}
        \caption{Qualitative retrieval results of SEAM Match-RCNN for the Regular-MovingFashion dataset. On the left, we show 3 frames sampled from the 10 frames used for aggregation. On the right the shop images retrieved starting from the closest match (left). The correct matches are represented with a green border.
        }
        \label{fig:retrievals2}
        \vspace{-1em}
    \end{figure*}

On Fig.~\ref{fig:retrievals2} results of SEAM Match-RCNN for the Regular-MovingFashion dataset are shown. Here, on street frames which exhibit more regularities, the shop items are vice versa more insidious than the TikTok ones, since they exhibit a lower variability, see for example the black female dresses of row 6. The same rationale holds for the white shirts and the black paints.

\begin{figure*}[t!]
        \centering
        \includegraphics[width=\linewidth]{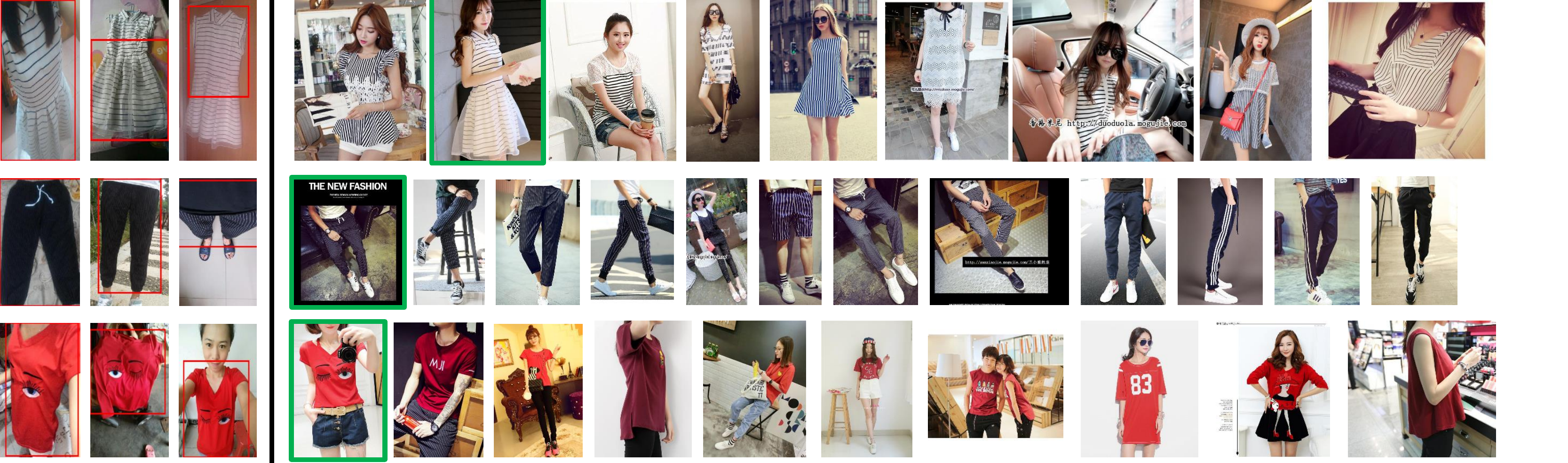}
        \caption{Qualitative retrieval results of SEAM Match-RCNN for the MultiDeepFashion2 dataset. On the left, we show 3 frames sampled from the 10 frames used for aggregation. On the right the shop images retrieved starting from the closest match (left). The correct matches are represented with a green border.   }
        \label{fig:retrievals3}
        \vspace{-1em}
    \end{figure*}

Finally, on  Fig.~\ref{fig:retrievals3} retrieval results on MultiDeepFashion2 are shown. Looking at the retrieval results, one can notice that shop items are way less regular/neutral than the ones on the MovingFashion (which anyway represent a more genuine excerpt of an e-commerce website): at the same time, street frames are often zoomed captures of the object of interest, in general offering a retrieval challenge different than the one on MovingFashion. The strong results obtained by SEAM Match-RCNN prove its versatility in working on a broader set of scenarios.

\section{Future perspectives}   \label{sec:add_conclusion} 
With SEAM Match-RCNN we showed how the contribution of multiple frames can boost the retrieval accuracy by 33\% on MultiDeepFashion2 w.r.t Single-frame approaches and by 69\% on the MovingFashion dataset. We also obtained new, state-of-the-art results on all of the benchmarks. Still, much progress has to be made in order to present a new product to the market: looking at the results, the probability of finding the correct shop match within the top 20 ranked shop images is 87\% on TikTok/Instagram videos. In order to connect all the dots available within the data, one has to exploit all of the details of the clothing items shown in some of the frames, something which we are currently not able to perform (in fact, we are discarding them with low attention), because they cannot be mapped to the general layout of the clothing item. Therefore, we should probably consider 3D atlases and have a common reference there.    

This setup can be attractive for many scenarios, for example: 1) a \emph{casual user} can match a video snippet of a nice outfit he/she has captured with a gallery of products (e.g. Zalando, Amazon, etc.); 2) a \emph{fast fashion company} can measure the similarity of clothing items contained in a viral video, or fashion show, with the items of its catalogue, deciding which item to promote the most; 3) Youtube videos can be automatically processed by \emph{video sharing platforms} to build valuable statistics of popular outfits and discover emerging trends.


\end{document}